\documentclass[runningheads]{llncs}
\usepackage{xcolor}

\usepackage{eccv}
\usepackage{array}

\usepackage{eccvabbrv}
\usepackage{placeins}
\usepackage{graphicx}
\usepackage{booktabs}
\usepackage{amsmath,amssymb,mathtools}
\usepackage{enumitem}
\usepackage{xcolor}
\usepackage{amssymb}
\usepackage[accsupp]{axessibility}  

\usepackage{hyperref}
\usepackage{multirow}
\usepackage{tabularx}
\usepackage{array}
\usepackage{ragged2e}
\usepackage{microtype}
\usepackage{booktabs}
\usepackage[table]{xcolor}
\newcolumntype{Y}{>{\centering\arraybackslash}X}
\newcolumntype{L}{>{\RaggedRight\arraybackslash}X}
\newcolumntype{P}[1]{>{\RaggedRight\arraybackslash}p{#1}}
\setlist[itemize]{leftmargin=1.4em,topsep=2pt,itemsep=1pt,parsep=0pt}
\setlist[enumerate]{leftmargin=1.6em,topsep=2pt,itemsep=1pt,parsep=0pt}
\newcommand{\code}[1]{\path{#1}}
\newcommand{\tablegroup}[1]{\rowcolor{gray!12}\multicolumn{3}{l}{\textbf{#1}}\\}
\begin{document}
\newcommand{\HN}[1]{\textcolor{red}{[\textbf{HN:} #1]}}
\newcommand{\yoshi}[1]{\textcolor{cyan}{[\textbf{Yoshi:} #1]}}

\newcommand{\cmark}{\textcolor{green!60!black}{\checkmark}}
\newcommand{\xmark}{\textcolor{red!70!black}{$\times$}}
\title{Feed-forward Motion In-betweening for Any 4D}

\author{
Hiroki Nishizawa\inst{1,2} \and
Hubert P. H. Shum\inst{3} \and
Yoshihiro Fukuhara\inst{1,2} \and
Hirokatsu Kataoka \inst{2} \and 
Shigeo Morishima\inst{1}
}

\authorrunning{H. Nishizawa et al.}

\institute{
\textsuperscript{1}Waseda University \quad
\textsuperscript{2}AIST \quad
\textsuperscript{3}Durham University
}
\maketitle
\begin{figure}
    \centering
    \includegraphics[width=0.99\linewidth]{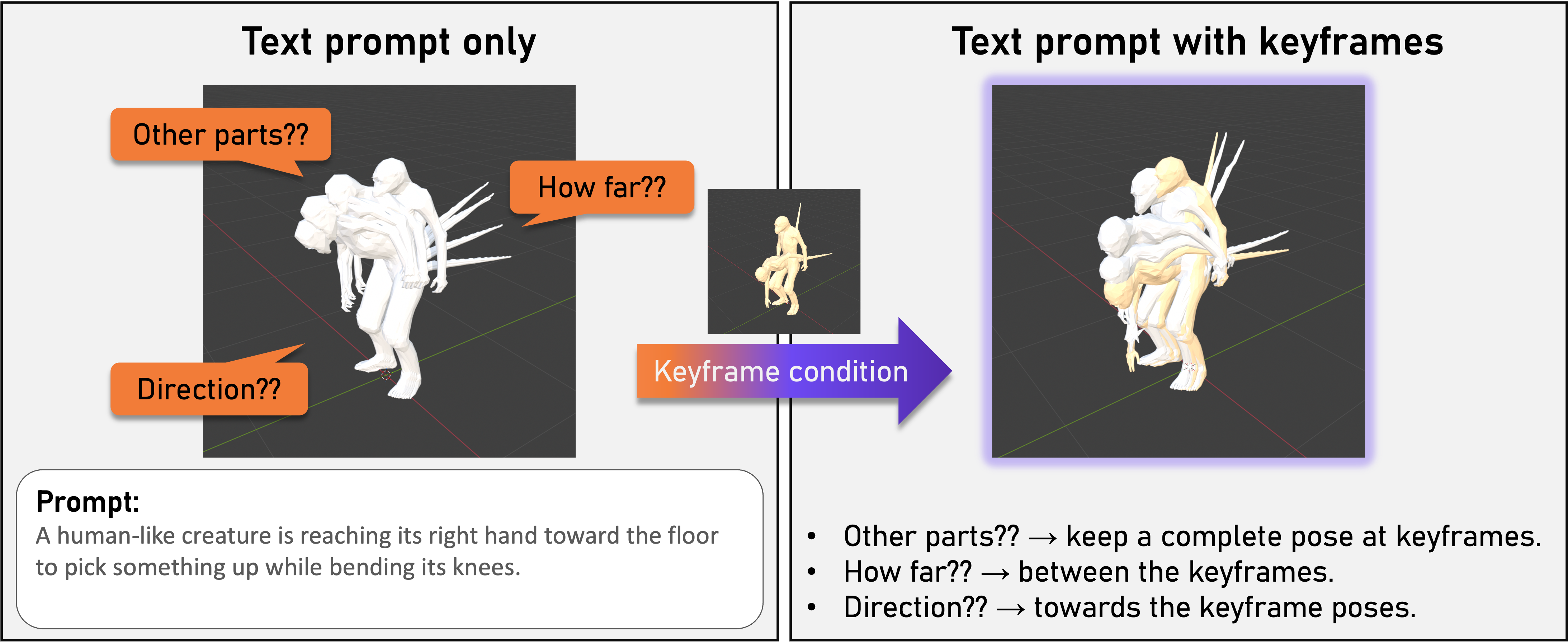}
    \caption{We introduce CompletionAny4D, the first feedforward motion in-betweenings for any 4D mesh, completes the 4D scenes, objects, animals, and characters from sparse keyframes and text prompts in a few seconds. }
    \label{fig:teaser}
\end{figure}
\begin{abstract}
4D dynamics (3D geometry evolving over time) is a fundamental representation of the physical world and plays a crucial role in world modeling (e.g., animation and games). Owing to the scarcity of large-scale, long-horizon 4D mesh data with arbitrary shapes, early text-to-4D methods rely on distillation or test-time optimization from video diffusion priors, making inference prohibitively slow. Recent feed-forward generators greatly reduce inference cost but offer limited spatiotemporal controllability, and short-horizon generation often leads to error accumulation in long-horizon sequences. We propose a novel feed-forward in-betweening framework for arbitrary 4D meshes with keyframe conditioning. Building on universal mesh-animation latents, we introduce a frame-wise mesh VAE that encodes each frame into topology-agnostic latent tokens anchored by a reference mesh for keyframe conditioning. We further introduce a keyframe-conditioned rectified flow model with an MMDiT backbone that synthesizes non-keyframe frames conditioned on sparse keyframes. Experiments show strong performance and improved controllability on both DyMesh16 and DyMesh32 benchmarks.
\keywords{4D generation \and mesh animation \and motion in-betweening \and rectified flow}
\end{abstract}
\section{Introduction}
Many real-world physical objects, such as humans, animals, plants, and everyday objects, undergo continuous deformation over time~\cite{Lyu2026CHORD}.
Motion in-betweening is a traditional technique in the animation process that generates natural and smooth transitions between keyframes~\cite{Harvey2020RobustMIB,cohan2024condmdi}. This approach provides users with the precise control needed to reproduce such real-world physical deformations. 

Despite its importance, generative modeling of 4D content remains challenging.
A major bottleneck is the scarcity of large-scale, long-horizon 4D datasets, especially for dynamic meshes with arbitrary topology~\cite{Lyu2026CHORD,Wu_2025_ICCV}.
Early progress has therefore relied on distillation or test-time optimization from powerful video diffusion priors~\cite{pmlr-v202-singer23a,bahmani2024fourdfy,ling2024aligngaussians,wu2025cat4d,Lyu2026CHORD}.
These approaches are attractive as they do not require curated 4D training data, but they typically demand expensive per-instance optimization.
For example, despite Choreographing a world can generate highly natural 4D scenes, it takes approximately 20 hours on an NVIDIA H200 GPU for just 41 frames~\cite{Lyu2026CHORD}, which severely limits practical deployment and interactive use. More recently, several works have shifted towards collecting/constructing dedicated 4D datasets and training feed-forward generators, achieving orders-of-magnitude faster synthesis at inference time~\cite{Wu_2025_ICCV,Li2025SS4D,ren2024l4gm}.

While existing models can produce visually plausible and natural motion, they introduce new limitations that hinder their use as controllable 4D content creation tools. 
First, because these models are fundamentally generative, the motion space they sample from can be highly diverse, making it difficult to obtain a specific desired motion.
In real-world applications, users often require spatiotemporal control (e.g., "raise the left arm with a slight bend at the elbow," "jump up and land on the platform," or "gradually bend the branch under increasing load"), which is hard to represent users' thoughts just by text.
Sparse keyframes have emerged as an effective control signal in motion synthesis~\cite{cohan2024condmdi,Wei2024DiffKFC,Bae_2025_ICCV}, and analogous sparse-conditioning paradigms have been explored for long-horizon video completion~\cite{Harvey2022FlexibleDiffusionLongVideos}.
Second, due to the lack of long, large-scale 4D training sequences, current methods often generate only short clips~\cite{Wu_2025_ICCV}.
A seemingly straightforward remedy is autoregressive generation; however, without explicit constraints, errors accumulate, and the motion drifts, making it difficult to maintain spatiotemporal consistency over long horizons~\cite{Zhu2025AR4D}.

To address the issues, we propose CompletionAny4D, a text and keyframe conditioned feedforward 4D generation framework.
The framework is based upon the idea to separate topology/shape information from motion information, which has shown to yield a robust latent representation for universal mesh animation \cite{Wu_2025_ICCV}. On top of this, we introduce two purposefully-built designs. First, we convert the sequence-wise DyMeshVAE into a frame-wise VAE that encodes each frame independently while preserving a shared shape anchor, preparing the system for keyframe conditioning. Second, we introduce keyframe-conditioned generation in the latent space, turning the problem into a motion-in-betweening task where non-keyframe frames are synthesized to satisfy spatiotemporal constraints named keyframes.
Crucially, the model is always conditioned on a reference mesh at $t=0$, which anchors topology-related factors, so the task is not purely text-driven but explicitly text-and-mesh-driven.
Finally, we combine this controllable latent representation with autoregressive generation under keyframe constraints, enabling long-horizon synthesis with substantially improved spatiotemporal control.
Our main contributions are as follows:
\renewcommand{\labelitemi}{\ensuremath{\cdot}}
\begin{itemize}
    \item \textbf{CompletionAny4D.}
    We propose CompletionAny4D, a text and keyframe conditioned feedforward framework that generates motion in-betweening mesh sequences for arbitrary-topology 4D meshes, enabling precise spatiotemporal controlling and consistent longer sequence generation while trained on short-horizon datasets.

    \item \textbf{Frame-wise mesh VAE.}
    We design a frame-wise mesh VAE that produces mesh topology latents and vertex trajectory latent for each frames anchored by a shared reference mesh, enabling direct keyframe latent injecting or replacements for motion in-betweening.

    \item \textbf{Keyframe-conditioned rectified flow.}
    We introduce a keyframe-conditioned Rectified Flow model~\cite{Liu2022RectifiedFlow} that generate each frame trajectory latents from text and keyframe latents with keyframe mask, enabling spatiotemporal controls via keyframe conditioning.

    \item \textbf{Long-horizon generation with sparse keyframe.}
    We further experiment with long-horizon synthesis via motion-in-betweening from sparse keyframes, producing long and spatiotemporally controllable 4D sequences while trained on short horizon.
\end{itemize}

\section{Related Work}
\subsection{4D Generation}
Early text-driven 4D generation extends score distillation sampling (SDS) to the spatiotemporal domain, distilling dynamic NeRF or Gaussian representations from pretrained diffusion priors. While avoiding paired 4D supervision, these approaches rely on expensive per-scene optimization and often suffer from spatiotemporal artifacts \cite{Wu2024SC4D,jiang2024consistent4d,bahmani2024fourdfy,animate124,ren2023dreamgaussian4d,alignyg,xian2023mav3d,4dgen,efficient4d,zeng2024stag4d}.
Another line improves controllability by synthesizing camera-time-conditioned multi-view videos and reconstructing 4D representations such as deformable 3D Gaussians \cite{wu2025cat4d,jiang2024animate3d,zhang2024fourdiffusion,watson2024fourdim,wang2024vidu4d,wu2024fourdgs}. However, these methods still require reconstruction-time optimization and depend on synchronized multi-view dynamic data.
Recent work instead amortizes 4D reconstruction and animation with feed-forward models, spanning large-scale 4D Gaussian reconstruction from monocular video to text-driven mesh animation \cite{tang2024lgm,ren2024l4gm,wang2024vidu4d,Wu_2025_ICCV}. Despite this progress, existing 4D datasets remain limited and dominated by short clips, making long-duration temporally coherent 4D generation challenging \cite{li2021fourdcomplete,mahmood2019amass,deitke2022objaverse,deitke2023objaversexl,wu2025cat4d,ren2024l4gm,Wu_2025_ICCV}.

\subsection{Motion In-betweening in 4D}
\subsubsection{Human Motion In-betweening.}
Motion in-betweening aims to synthesize dense motion from sparse keyframes. Early learning-based methods explored recurrent, adversarial, and structured latent-space formulations to improve robustness and controllability~\cite{Harvey2020RobustMIB,Kaufmann2020MotionInfilling,Oreshkin2022DeltaInterpolator,Qin2022TwoStageTransformers,Kim2022ConditionalMIB}. More recent approaches build on strong diffusion priors for human motion generation~\cite{tevet2022mdm,zhang2022motiondiffuse,karunratanakul2023gmd} and formulate in-betweening as sparse-conditioned denoising, infilling, or masked motion completion~\cite{cohan2024condmdi,Wei2024DiffKFC,Xie2024OmniControl,watanabe2026projflow,guo2023momask,geng2024keymotion}. Several extensions further consider long-horizon transitions and practical animation settings such as intermediate keyframe prediction, imprecise timing, and coarse blocking poses~\cite{Qin2024RobustDiffusionMIB,Hong2024LongTermMIB,Zheng2025AutoKeyframe,Goel2025ImpreciseTiming,Goel2025BlockingPoses}. However, these methods are largely built on fixed skeletal parameterizations and do not naturally extend to dynamic meshes with arbitrary and time-varying topology.
\subsubsection{Motion In-betweening for Arbitrary Shapes.}
Moving beyond human skeleton motion, in-betweening must cope with diverse morphology, limited per-character data, and the absence of a unified rig.
AnyMoLe~\cite{yun2025anymole} tackles this by leveraging video diffusion priors and character adaptation, enabling motion in-betweening for unseen characters without requiring character-specific motion datasets.
Related work in the image/video domain adapts large image-to-video diffusion models for keyframe interpolation~\cite{Wang2025GenerativeInbetweening}, and recent approaches combine character-agnostic motion generation with physics-based adaptation to scale across characters with different morphology~\cite{Qin2025ScalableMIB}.
For general 4D generation, sparse control signals (e.g., sparse views, trajectories, or frames) have been explored for dynamic reconstruction and motion transfer~\cite{Wu2024SC4D,Li2024DreamMesh4D,Bahmani2024TC4D,Nag2025In2_4D}, but they typically do not provide constraints over mesh keyframes under arbitrary topology.
To compare these methods, our proposed method enables explicit keyframe conditioning and generate arbitrary topology mesh motion-in-betweenings as a feedforward.

\section{Method}
\subsection{Problem Setup}
We consider text-and-mesh-driven motion in-betweening with optional sparse keyframes.
Let $\{\mathcal{M}_t\}_{t=0}^{T}$ denote a dynamic mesh sequence, where each frame $\mathcal{M}_t$ is the same topology in the sequence.
To account for topology-related factors, we always condition the model on a reference mesh
$\mathcal{M}_0$ (the $0$-th frame), which is included during both training and inference.
The inputs are:
(i) a text prompt $\mathbf{p}$,
(ii) a mandatory reference mesh $\mathcal{M}_0$ (mesh/topology conditioning), and
(iii) start and end keyframes and optional set of additional user-provided keyframes 
$\mathcal{K}=\{(1,\mathcal{M}_{1})\,(T,\mathcal{M}_{T})\,(t_i,\mathcal{M}_{t_i})\}_{i=1}^{|\mathcal{K}|}$ with $t_i\in\{2,\dots,T-1\}$.
Our goal is to synthesize the missing (in-between) frames and produce a full mesh sequence
$\{\widehat{\mathcal{M}}_t\}_{t=0}^{T}$ such that
(a) the overall motion and appearance are semantically aligned with $\mathbf{p}$, and
(b) the generated frames are spatiotemporally consistent with the conditioned meshes,
while keeping them fixed:
$\widehat{\mathcal{M}}_0=\mathcal{M}_0$ and
$\widehat{\mathcal{M}}_{t}=\mathcal{M}_{t}$ for all $(t,\mathcal{M}_{t})\in\mathcal{K}$.
\begin{figure}
    \centering
    \includegraphics[width=0.99\linewidth]{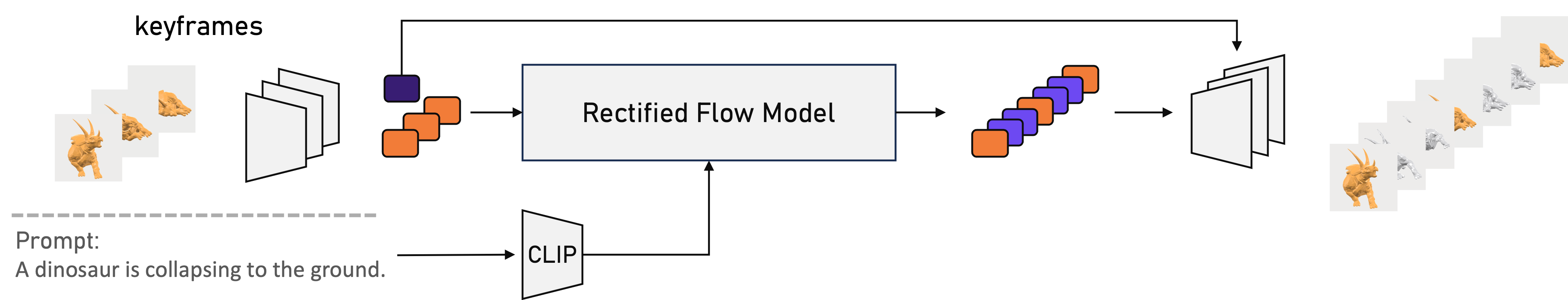}
    \caption{Method overview of our CompletionAny4D, a framework of frame-wise latent encoding and generation for motion in-betweening.}
    \label{fig:overview}
\end{figure}
\subsection{The CompletionAny4D Framework}
The CompletionAny4D mesh in-betweening pipeline is illustrated in Fig.~\ref{fig:overview}.
Given a canonical mesh $\mathcal{M}_0$ and sparse keyframe meshes $\{(t,\mathcal{M}_{t})\}_{t\neq0}$, we first encode $\mathcal{M}_{0}$ and other keyframes $\{(t,\mathcal{M}_{t})\}_{t\neq0}$ to obtain a shared shape structure latent $\widehat{\mathbf{V}}^n_0$ and per-frame trajectory latents $\widehat{\mathbf{V}}^n_t$.
To condition the rectified flow model, we construct a latent conditioning sequence by injecting clean latents only at keyframe timesteps and using noised placeholders elsewhere.

Concretely, with a keyframe mask $m_t\in\{0,1\}$, we form the noisy latent condition with timestep~$\tau$:
\begin{equation}
\widetilde{\mathbf{V}}_t^{noised} \;=\; m_t\,\mathbf{V}^n_t \;+\; (1-m_t)\,\mathbf{V}^{\mathrm{n}}_t(\tau),
\end{equation}
and feed $\widetilde{\mathbf{V}}=\{\widetilde{\mathbf{V}}^n_t\}$ (together with $\widehat{\mathbf{V}}^n_0$ and the text prompt $\textit{\textbf{p}}$) as the conditioning input to the rectified flow model.
We then sample the flow and integrate the corresponding ODE to obtain non-keyframe latents $\{\widehat{\mathbf{V}}^n_t\}$, which are decoded jointly with $\widehat{\mathbf{V}^n_0}$ to reconstruct the full 4D mesh sequence.

Sparse keyframe conditioning requires a frame-aligned latent space, where each latent token has a one-to-one correspondence with a frame. Such a representation enables direct clamping of keyframe latents without implicitly modifying non-keyframe timesteps. Therefore, we need a frame-wise VAE encodes frames independently, preserving frame-level correspondence and enabling straightforward, localized keyframe conditioning. This is in contrast with existing work such as \cite{Wu_2025_ICCV}, which temporally compresses motion trajectories during encoding, which couples information across frames and hinders explicit sparse keyframe control. 

\begin{figure}
    \centering
    \includegraphics[width=0.99\linewidth]{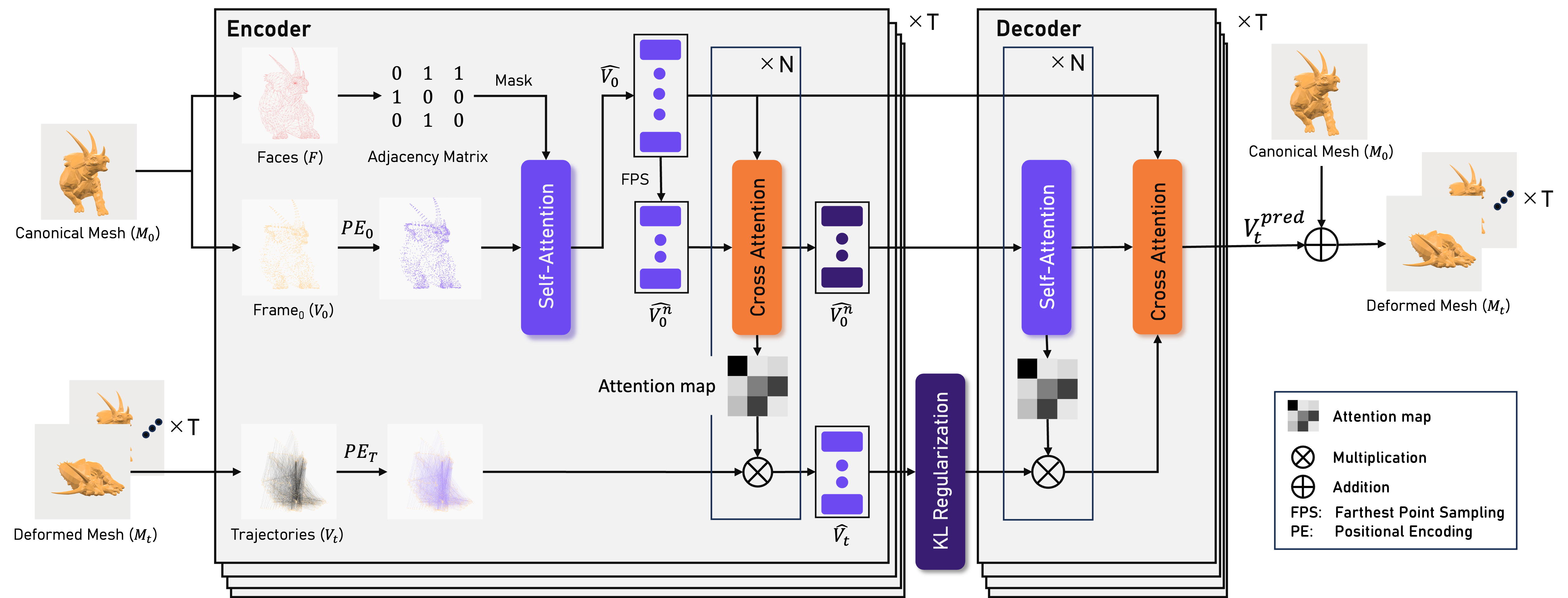}
    \caption{The architecture of our frame-wise mesh VAE}
    \label{fig:placeholder}
\end{figure}
\subsection{Frame-wise VAE}

To enable keyframe conditioning, we require a time-aligned latent representation where each frame index corresponds to an explicit latent code that can be clamped at keyframes.
In contrast, existing work such as \cite{Wu_2025_ICCV} encodes an entire clip into a fixed-length token set, effectively compressing the temporal dimension; while suitable for unconditional generation, this design is inconvenient for imposing hard constraints at specific frames.

In particular, our proposed frame-wise mesh VAE encodes meshes independently per frame while disentangling shape structure from motion.
Given a reference mesh $\mathcal{M}_0$, we extract a shape structure token sequence $\widehat{\mathbf{V}}^n_0$ that is shared across the entire sequence.
For each timestep $t$, we encode the mesh frame $\mathcal{M}_t$ into a per-frame motion token sequence $\widehat{\mathbf{V}}_t$ and we do FPS(Farthest Point Sampling) to reduce the vertex counts to n. Our latents is consisted of full vertex shape structure latent $\widehat{\mathbf{V}}_0$ and reduced vertex shape structure latent $\widehat{\mathbf{V}}^n_0$ and reduced vertex trajectory latent $\widehat{\mathbf{V}}^n_t$.
A shared decoder reconstructs each frame from $\widehat{\mathbf{V}}_0, \widehat{\mathbf{V}}^n_0, \widehat{\mathbf{V}}^n_t$.

\subsection{Keyframe-conditioned Frame-wise Rectified Flow Model}
\subsubsection{Rectified Flow.}
Given a data sample $\mathbf{x}_1 \sim \pi_1$ and a noise sample $\mathbf{x}_0 \sim \pi_0$ (typically $\mathcal{N}(0,\mathbf{I})$), rectified flow defines a straight-line path to learn the transport dynamics between a base distribution and a data distribution by learning an ODE velocity field~\cite{Liu2022RectifiedFlow}:
\begin{equation}
\mathbf{x}_t = (1-t)\,\mathbf{x}_0 + t\,\mathbf{x}_1, \quad t\in[0,1].
\end{equation}
The target velocity along this path is constant:
\begin{equation}
\mathbf{\upsilon}_t(\mathbf{x}_t,t) = \frac{\mathrm{d}\mathbf{x}_t}{\mathrm{d}t} = \mathbf{x}_1 - \mathbf{x}_0.
\end{equation}
During training, $\mathbf{\upsilon}_\theta(\mathbf{x}_t,t,\mathbf{c})$ is trained to regress $\mathbf{\upsilon}_t$ under (optional) conditioning $\mathbf{c}$.
In practice, we sample $t\sim\mathcal{U}(0,1)$, form $\mathbf{x}_t=(1-t)\mathbf{x}_0+t\mathbf{x}_1$, and minimize the squared regression loss:
\begin{equation}
\mathcal{L}_{\mathrm{RF}}(\theta)
=
\mathbb{E}_{\mathbf{x}_1\sim\pi_1,\;\mathbf{x}_0\sim\pi_0,\;t\sim\mathcal{U}(0,1)}
\Big[
\big\|
\mathbf{\upsilon}_\theta(\mathbf{x}_t,t,\mathbf{c})-\mathbf{\upsilon}_t(\mathbf{x}_t,t)
\big\|_2^2
\Big].
\end{equation}
At inference, samples are obtained by solving the learned ODE
\begin{equation}
\frac{\mathrm{d}\mathbf{x}}{\mathrm{d}t} = \mathbf{\upsilon}_\theta(\mathbf{x},t,\mathbf{c}),
\quad \mathbf{x}(0)=\mathbf{x}_0,\ \mathbf{x}_0\sim\pi_0,
\end{equation}
and reading out $\mathbf{x}(1)$.
Using a $K$-step discretization, this corresponds to:
\begin{equation}
\mathbf{x}_{t_{k+1}}=\mathbf{x}_{t_k}+\Delta t\;\mathbf{\upsilon}_\theta(\mathbf{x}_{t_k},t_k,\mathbf{c}),
\quad t_k=\frac{k}{K},\ \Delta t=\frac{1}{K}.
\end{equation}
Rectified flow has been shown to scale well with transformer backbones and to support multimodal conditioning mechanisms (\eg, MMDiT)~\cite{Esser2024ScalingRectifiedFlowTransformers}.

\subsubsection{The Frame-wise Rectified Flow Model.}
We model the distribution of frame-wise trajectory latent sequences
$\widehat{\mathbf{V}}^n = [\widehat{\mathbf{V}}^n_1,\dots,\widehat{\mathbf{V}}^n_T]$
with a conditional rectified flow.
Let $\epsilon \sim \mathcal{N}(0,\mathbf{I})$ be a base latent with the same shape as $\widehat{\mathbf{V}}^n$.
Following rectified flow, we define the straight-line interpolation:
\begin{equation}
\widehat{\mathbf{V}}^n(t) = (1-t)\,\epsilon + t\,\widehat{\mathbf{V}}^n,\quad
t\in[0,1].
\label{formula:noised_traj_latent}
\end{equation}
The corresponding target velocity is again constant:
\begin{equation}
\mathbf{\upsilon}_t(\widehat{\mathbf{V}}^n(t),t) =\frac{\mathrm{d}\widehat{\mathbf{V}}^n(t)}{\mathrm{d}t}=\widehat{\mathbf{V}}^n-\epsilon.
\end{equation}
The conditioning signal $\mathbf{c}$ fuses the text prompt \textbf{\textit{p}}, the shape structure latent $\widehat{\mathbf{V}}^n_0$, and the noised trajectory latent $\widehat{\mathbf{V}}^n(t)$.
We adopt an MMDiT-style transformer backbone~\cite{Esser2024ScalingRectifiedFlowTransformers,Peebles_2023_ICCV} for flexible multi-modal fusion, and parameterize the velocity field as:
\begin{equation}
\mathbf{\upsilon}_\theta(\widehat{\mathbf{V}}^n(t),t,\mathbf{c}^n)
=
\mathrm{MMDiT}(\widehat{\mathbf{V}}^n(t),\widehat{\mathbf{V}}^n_0,\textbf{\textit{p}},t).
\end{equation}
During training, by instantiating the rectified flow objective for trajectory latents, we train:$\mathbf{\upsilon}_\theta$ with:
\begin{equation}
\mathcal{L}_{\mathrm{frameRF}}(\theta)
=
\mathbb{E}_{\widehat{\mathbf{V}}^n,\;\epsilon\sim\mathcal{N}(0,\mathbf{I}),\;t\sim\mathcal{U}(0,1)}
\Big[
\big\|
\mathbf{\upsilon}_\theta(\widehat{\mathbf{V}}^n(t),t,\mathbf{c}^n)-\mathbf{\upsilon}_t(\widehat{\mathbf{V}}^n(t),t)
\big\|_2^2
\Big].
\label{formula:loss_rf}
\end{equation}

During sampling given a prompt \textbf{\textit{p}} and shape structure latent $\widehat{\mathbf{V}}^n_0$, we sample $\epsilon\sim\mathcal{N}(0,\mathbf{I})$ and solve:
\begin{equation}
\frac{\mathrm{d}\widehat{\mathbf{V}}^n(t)}{\mathrm{d}t}
=
\mathbf{\upsilon}_\theta(\widehat{\mathbf{V}}^n(t),t,\mathbf{c}),
\quad
\widehat{\mathbf{V}}^n(0)=\epsilon,
\label{formula:ode_solving_framewise}
\end{equation}
and take $\widehat{\mathbf{V}}^n(1)$ as the generated trajectory latent sequence.
\begin{figure}
    \centering
    \includegraphics[width=0.99\linewidth]{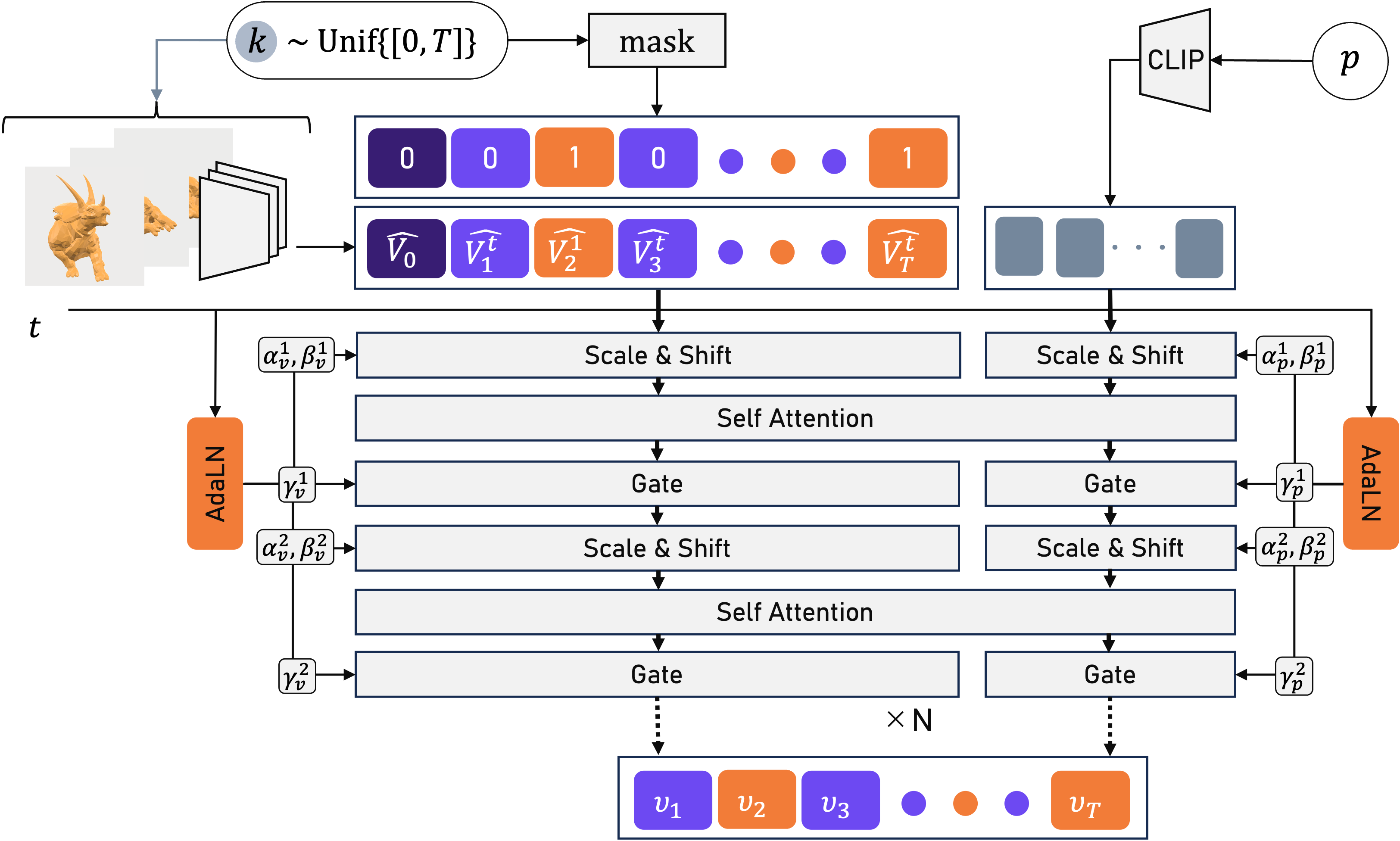}
    \caption{The architecture of our Keyframe-conditioned frame-wise Rectified Flow model.}
    \label{fig:method-overview}
\end{figure}
\subsubsection{Baseline: Keyframe Replacement at Sampling-time.}
One simple keyframe conditioning for RFmodel is to replace the noisy latents that correspond to the keyframes with the noisy latents computed from keyframe meshes for each ODE solving step. We set the frame-wise Rectified Flow model with keyframe replacement at sampling-time as our baseline for the in-betweening method.

From our definition of the noisy trajectory latents Eq.~\ref{formula:noised_traj_latent} and the ODE solving step Eq.~\ref{formula:ode_solving_framewise}, we can formulate this replacement system as:
\begin{equation}
\begin{aligned}
\widetilde{\mathbf{V}}^{n}(\tau_k)
&= (\mathbf{1}-\mathbf{m})\odot \widehat{\mathbf{V}}^{n}(\tau_k)
\;+\; \mathbf{m}\odot \widehat{\mathbf{V}}^{n}_{\mathrm{t}}(\tau_k), ~t\sim\mathcal{K}\\[2pt]
\widehat{\mathbf{V}}^{n}(\tau_{k+1})
&= \widehat{\mathbf{V}}^{n}(\tau_k)
\;+\; \Delta \tau\;
\mathbf{\upsilon}_{\theta}\!\left(\widetilde{\mathbf{V}}^{n}(\tau_k),\, \tau_k,\, \mathbf{c}\right), ~ \tau_k=\frac{k}{K},\ \Delta \tau=\frac{1}{K}.
\end{aligned}
\label{formula:keyframe_replacing}
\end{equation}

\subsubsection{Proposed Keyframe Encoding and Masking.}
Replacing keyframes only at sampling time provides weak motion control and often yields unnatural motions. We therefore develop a CondMDI-style masked latent replacement scheme for our frame-wise rectified flow model. 

Let $\mathbf{m}\in\{0,1\}^{T}$ be a binary mask over frame indices $i\in\{1,\dots,T\}$, where $m_i=1$ denotes an observed keyframe. To account for the shape token, we prepend a zero and define the extended mask $\widehat{\mathbf{m}} := [0;\mathbf{m}] \in \{0,1\}^{T+1}$, whose leading zero corresponds to the shape structure latent $\widehat{\mathbf{V}}_0^n$. Let $\widehat{\mathbf{V}}_{\mathrm{key}}^n(1)$ denote the trajectory-latent sequence aligned with the selected keyframes, i.e., its entries at constrained frames are the corresponding clean keyframe latents. At rectified-flow time $\tau$, the conditioning sequence fed to MMDiT is:
\begin{equation}
\tilde{\mathbf{Z}}(\tau)
=
\mathrm{cat}\!\Big(
\widehat{\mathbf{V}}_0^n,\,
\mathbf{m}\odot \widehat{\mathbf{V}}_{\mathrm{key}}^n(1)
+
(\mathbf{1}-\mathbf{m})\odot \widehat{\mathbf{V}}^n(\tau)
\Big),
\label{formula:mask_concat}
\end{equation}
where $\odot$ denotes element-wise multiplication with $\mathbf{m}$ broadcast along the latent dimension. The resulting conditioned velocity field is:
\begin{equation}
\mathbf{\upsilon}_\theta(\tilde{\mathbf{Z}}(\tau),\tau)
=
\mathrm{MMDiT}(\widehat{\mathbf{m}},\tilde{\mathbf{Z}}(\tau),\textbf{\textit{p}},\tau).
\end{equation}

During training, we first sample the number of keyframes as $k \sim \mathrm{Unif}\{0,\dots,5\}$ and then uniformly choose a subset $\mathcal{K}\subset\{1,\dots,T\}$ with $|\mathcal{K}|=k$. Since keyframe masking only changes the conditioning input, the training objective remains identical to the frame-wise rectified flow loss in Eq.~\eqref{formula:loss_rf}.

At inference, given the prompt \textbf{\textit{p}}, the shape structure latent $\widehat{\mathbf{V}}_0^n$, and the observed keyframe latents, we solve Eq.~\eqref{formula:ode_solving_framewise} using the masked conditioning sequence $\tilde{\mathbf{Z}}(\tau)$.

\section{Experiments}
\subsection{Experiment Settings}
\subsubsection{Dataset Curation.}
We use DyMesh16, the 16-frame subset of the large-scale DyMesh dataset introduced by AnimateAnyMesh~\cite{Wu_2025_ICCV}.
DyMesh is curated from diverse sources and contains dynamic mesh sequences with arbitrary topology; DyMesh16 comprises millions of 16-frame clips, making it suitable for training feed-forward 4D generators.
For computational efficiency, we filter the data to meshes with 512--4,096 vertices and split the resulting set into an 80/20 train/validation split.
We train our frame-wise DyMeshVAE and Rectified Flow models on this subset.
We find that our frame-wise VAE achieves competitive reconstruction performance compared to a sequence-wise VAE trained on the same split, as well as the publicly released model trained on a larger dataset.
These results suggest that our curated subset is sufficient to reproduce the released model's reconstruction accuracy and to support our downstream experiments.

\subsubsection{Implementation Detail.}
All models are trained on DyMesh16 with the number of vertices restricted from 512 to 4096, resulting in approximately 260K samples in total(208K for training, 52K for validation).
For our frame-wise mesh VAE, we adopt the same architectural design as AnimateAnyMesh~\cite{Wu_2025_ICCV}, except for the input frame configuration and latent dimensionality. The encoder consists of 8 cross-attention layers, while the decoder is composed of 8 self-attention layers. In the encoder, latent tokens are sampled using FPS to 512 spatial tokens. The channel dimension is set to 12, yielding latent representations $\widehat{\mathbf{V}}^n_0$ and $\widehat{\mathbf{V}}^n_t$ of size $512 \times 12$.
The VAE is optimized using Adam~\cite{Kingma2015Adam} with a learning rate of $1\times10^{-4}$ for 200 epochs on 16 NVIDIA H200 GPUs. Our proposed frame-wise VAE differs from the 16-frame DyMeshVAE~\cite{Wu_2025_ICCV} only in the number of input mesh frames and the latent channel dimensionality; all other architectural components and training configurations remain identical.
The results reported in Table~\ref{tab:framewiseVAE} are obtained by training the VAE with a learning rate of $1\times10^{-4}$ for 1000 epochs on 16 NVIDIA H200 GPUs.
For the rectified flow (RF) models, we follow AnimateAnyMesh~\cite{Wu_2025_ICCV} and employ a pretrained CLIP ViT-L/14~\cite{Radford2021CLIP} as the text encoder, with a maximum sequence length of 77 tokens.
Both our Frame-wise RFmodel and Keyframe-conditioned RFmodel are trained using Adam~\cite{Kingma2015Adam} with a learning rate of $2\times10^{-4}$ for 1000 epochs on 32 NVIDIA H100 GPUs.

\subsubsection{Evaluation Metrics.}
We conduct both quantitative and qualitative evaluations to assess motion naturalness, accuracy and controllability. For motion accuracy, we report RMSE between generated meshes and ground-truth meshes, averaged over all vertices and non-keyframes.
Additionally, to evaluate the motion similarity and shape structure consistency, we compute Dynamic-Time Warping and Chamfer distance.
For motion naturalness, we follow the evaluation protocol of AnimateAnyMesh~\cite{Wu_2025_ICCV} and adopt VBench~\cite{VBench}, including I2V Subject Similarity, Motion Smoothness, and Aesthetic Quality (denoted as I2V, M.sm, and Aest.Q, respectively).
Despite AnimateAnyMesh~\cite{Wu_2025_ICCV} conduct only 10 sample evaluation, we evaluated the generated results for 100 samples for a higher reliable evaluation.
We further conduct a user study following AnimateAnyMesh~\cite{Wu_2025_ICCV} protocol. We recruited 20 participants and asked to rate the generated motions on a 5-point Likert scale along four criteria: (i) motion naturalness, (ii) text-motion alignment, (iii) shape preservation, and we newly added (iv) fidelity to the ground-truth motion (denoted as Natur, Text, Shape, and Control, respectively). 
Despite AnimateAnyMesh~\cite{Wu_2025_ICCV} released its model weights, the dataset split strategy did not provided, and we are not able to conduct an open evaluation. In each tables, there are marks in the "Open" columns indicate whether the method evaluated open setting or not.
Finally, we report inference speeds for meshes have 512, 1024, 2048, and 4096 vertices for 16 and 31 frames measured on a single NVIDIA H100 GPU.

\subsection{Comparisions}
\begin{table*}[t]
\centering
\caption{The short-horizon evaluation on 100 samples from DyMesh16. We use the first and last frames and randomly sampled K frames for the keyframe constraints. Vertex accuracy metrics (RMSE, Chamfer, and DTW) are scaled by $\times 100$ for readability.}
\label{tab:short_horizon}
{\scriptsize
\setlength{\tabcolsep}{2pt}
\renewcommand{\arraystretch}{1.05}

\begin{tabular*}{\textwidth}{@{}@{\extracolsep{\fill}}l c c ccc ccc@{}}
\toprule
\multirow{2}{*}{\textbf{Method}} &
\multirow{2}{*}{\textbf{Open}} &
\multirow{2}{*}{\textbf{K}} &
\multicolumn{3}{c}{\textbf{Vertex acc. ($\times100$)}} &
\multicolumn{3}{c}{\textbf{VBench}} \\
\cmidrule(lr){4-6}\cmidrule(lr){7-9}
& & &
\textbf{RMSE$\downarrow$} &
\textbf{Chamfer$\downarrow$} &
\textbf{DTW$\downarrow$} &
\textbf{I2V$\uparrow$} &
\textbf{M.sm$\uparrow$} &
\textbf{Aest.Q$\uparrow$} \\
\midrule
AnimateAnyMesh~\cite{Wu_2025_ICCV}
& \xmark & -   & 6.424 & 29.45  & 27.96  & \textbf{0.9831} & \textbf{0.9946} & \textbf{0.4626} \\
\midrule
\multirow{4}{*}{Baseline}
& \cmark & K=0 & 9.061 & 11.19  & 3.498 & 0.9031  & 0.9874  & 0.4161 \\
& \cmark & K=1 & 8.111 & 9.687 & 2.916 & 0.9033  & 0.9857  & 0.4167 \\
& \cmark & K=2 & 7.242 & 8.452 & 2.528 & 0.9017  & 0.9848  & 0.4166 \\
& \cmark & K=3 & 6.438 & 7.299 & 2.138 & 0.9036  & 0.98e45  & 0.4169 \\
\midrule
\multirow{4}{*}{\textbf{Ours}}
& \cmark & K=0 & 2.907 & 4.551 & 1.226  & 0.9157  & 0.9903  & 0.4164 \\
& \cmark & K=1 & 2.025 & 3.535 & 0.8718 & 0.9184  & 0.9905  & 0.4164 \\
& \cmark & K=2 & \underline{1.864} & \underline{3.387} & \underline{0.8037} & \underline{0.9189} & \underline{0.9905} & 0.4169 \\
& \cmark & K=3 & \textbf{1.852} & \textbf{3.377} & \textbf{0.7859} & \underline{0.9190} & \underline{0.9905} & \underline{0.4171} \\
\bottomrule
\end{tabular*}
}
\end{table*}

\begin{table*}[t]
\centering
\caption{The long-horizon evaluation on 100 samples from DyMesh32. 
The vertex accuracy metrics (RMSE, Chamfer, and DTW) are scaled by $\times 100$ for readability.}
\label{tab:long_horizon}
\scriptsize
\setlength{\tabcolsep}{2pt}
\renewcommand{\arraystretch}{1.05}

\begin{tabular*}{\textwidth}{@{\hspace{3pt}}@{\extracolsep{\fill}}l c c c c c c c@{\hspace{3pt}}}
\toprule
\textbf{Method} &
\textbf{Open} &
\multicolumn{3}{c}{\textbf{Vertex acc. ($\times100$)}} & 
\multicolumn{3}{c}{\textbf{VBench}} \\
\cmidrule(lr){3-5}\cmidrule(lr){6-8}
& & \textbf{RMSE$\downarrow$} &
\textbf{Chamfer$\downarrow$} &
\textbf{DTW$\downarrow$} &
\textbf{I2V$\uparrow$} &
\textbf{M.sm$\uparrow$} &
\textbf{Aest.Q$\uparrow$} \\
\midrule
AnimateAnyMesh~\cite{Wu_2025_ICCV} & \xmark & 15.63 & 15.85 & 7.672 & \textbf{0.9465} & \textbf{0.9969} & \underline{0.4527} \\
\midrule
Baseline   & \cmark & \underline{8.666} & \underline{10.46} & \underline{3.488} & 0.9076 & 0.9869 & 0.4581 \\
\textbf{Ours} & \cmark & \textbf{2.132} & \textbf{3.180} & \textbf{0.7272} & \underline{0.9322} & \underline{0.9951} & \textbf{0.4531} \\
\bottomrule
\end{tabular*}
\end{table*}

\begin{table*}[t]
\centering
\caption{User study results (5-point Likert scale). Left: DyMesh16 (short-horizon). Right: DyMesh32 (long-horizon).}
\label{tab:userstudy} 
\scriptsize
\setlength{\tabcolsep}{1.5pt} 
\renewcommand{\arraystretch}{1.0}
\definecolor{oursColor}{HTML}{4C9AFF}
\begin{tabular*}{\textwidth}{@{}@{\extracolsep{\fill}}l c c c c c @{\hspace{2pt}} c c c c@{}}
\toprule
\multirow{2}{*}{\textbf{Method}} &
\multirow{2}{*}{\textbf{Open}} &
\multicolumn{4}{c}{\shortstack[c]{\textbf{DyMesh16}}} &
\multicolumn{4}{c}{\shortstack[c]{\textbf{DyMesh32}}} \\
\cmidrule(lr){3-6}\cmidrule(lr){7-10}
& &
\textbf{Natur$\uparrow$} & \textbf{Text$\uparrow$} & \textbf{Shape$\uparrow$} & \textbf{Control$\uparrow$} &
\textbf{Natur$\uparrow$} & \textbf{Text$\uparrow$} & \textbf{Shape$\uparrow$} & \textbf{Control$\uparrow$} \\
\midrule
\shortstack[l]{AnimateAnyMesh\cite{Wu_2025_ICCV}} & \xmark
& \textbf{4.10} & 3.14 & \textbf{4.28} & 2.63
& \underline{3.22} & \underline{3.13} & \underline{3.46} & 1.96 \\
\midrule
Baseline & \cmark
& 2.84 & 2.97 & 2.50 & \underline{2.64}
& 2.61 & 2.83 & 2.61 & \underline{2.72} \\
\textbf{Ours} & \cmark
& \underline{3.74} & \textbf{3.37} & \underline{3.15} & \textbf{4.12}
& \textbf{3.85} & \textbf{3.33} & \textbf{3.50} & \textbf{4.13} \\
\bottomrule
\end{tabular*}
\end{table*}

\begin{table*}[t]
\centering
\caption{End-to-end inference time (seconds) from encoding to final decoding across vertex resolutions for 16-frame and 31-frame sequences.}
\label{tab:vertex_resolution}
\scriptsize
\setlength{\tabcolsep}{2pt}
\renewcommand{\arraystretch}{1.1}

\begin{tabular*}{\textwidth}{@{}@{\extracolsep{\fill}}l c c c c c c@{}}
\toprule
\textbf{Method} & \textbf{Frames} & \textbf{512} & \textbf{1024} & \textbf{2048} & \textbf{4096} & \textbf{Mean} \\
\midrule
\multirow{2}{*}{AnimateAnyMesh~\cite{Wu_2025_ICCV}}
& 16 & 2.936 & 2.987 & 3.236 & 3.474 & 3.158 \\
& 31 & 4.438 & 4.583 & 4.934 & 5.504 & 4.865 \\
\midrule
\multirow{2}{*}{Baseline}
& 16 & \underline{2.869} & \underline{2.907} & \underline{3.064} & \underline{3.175} & \underline{3.003} \\
& 31 & \underline{4.274} & \textbf{4.300} & \underline{4.664} & \underline{5.002} & \underline{4.560} \\
\midrule
\multirow{2}{*}{\textbf{Ours}}
& 16 & \textbf{2.716} & \textbf{2.721} & \textbf{2.860} & \textbf{2.921} & \textbf{2.805} \\
& 31 & \textbf{4.239} & \underline{4.357} & \textbf{4.663} & \textbf{4.677} & \textbf{4.484} \\
\bottomrule
\end{tabular*}
\end{table*}

\subsubsection{Keyframe conditioning for short horizons.}
We evaluate spatiotemporal controllability and motion quality on 16-frame motion in-betweening. As a baseline, we use the frame-wise Rectified Flow model with keyframe replacement only at sampling time. We use the first and last frames and randomly sampled K frames from other frames and used as keyframes.

\begin{figure}
    \centering
    \includegraphics[width=0.99\linewidth]{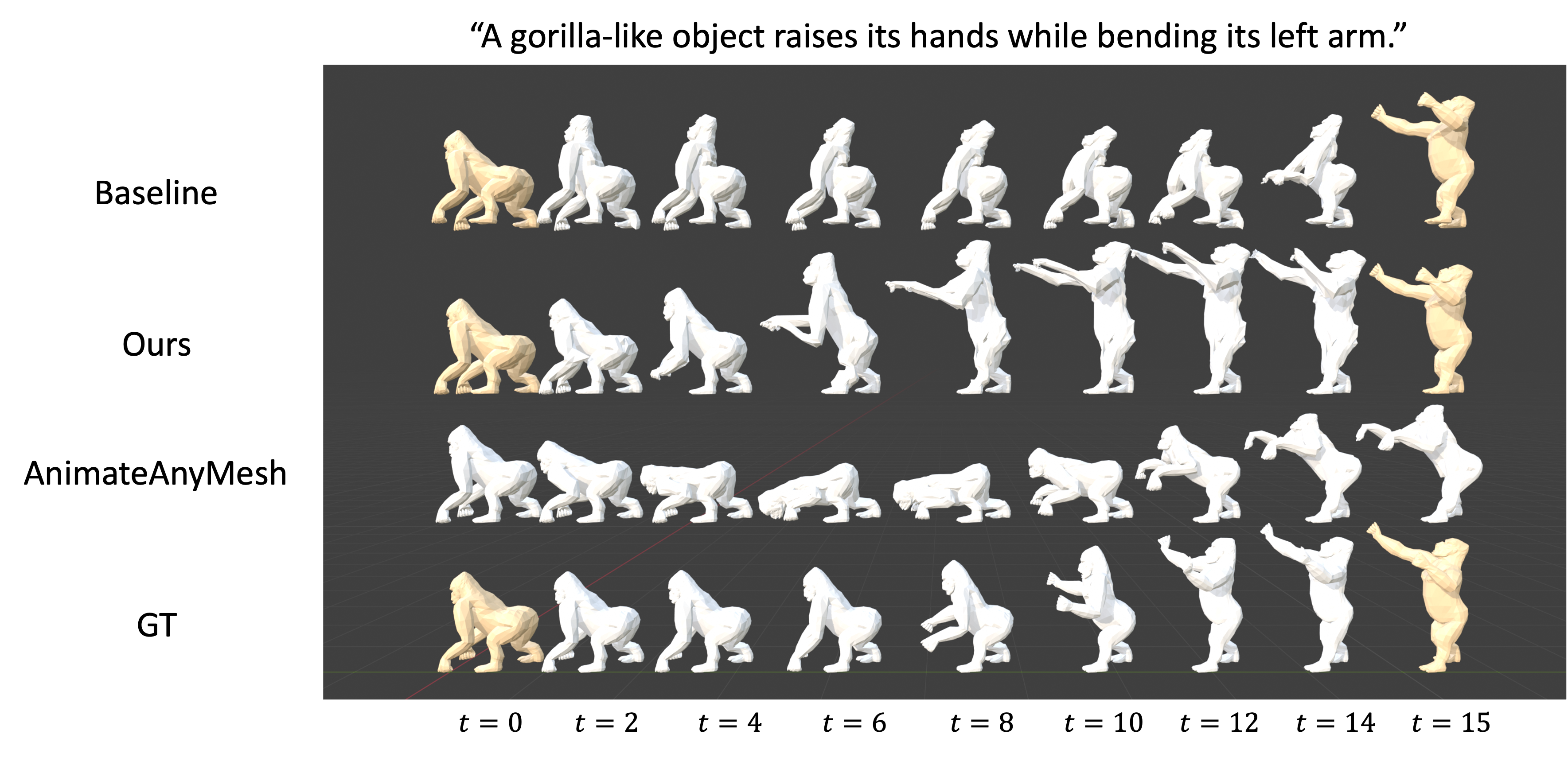}
    \caption{A qualitative result on Dymesh16 dataset. We sampled 7 frames from non-keyframes, the orange colored meshes are used as the keyframes which are conditioned for each methods.}
    \label{fig:main_16}
\end{figure}

Qualitative results are shown in Figure~\ref{fig:main_16}. AnimateAnyMesh~\cite{Wu_2025_ICCV} produces natural-looking motions, but struggles to satisfy the poses and motion cues specified by the text prompt, as illustrated in the top row of the figure. In contrast, our method generates smooth transitions between the start and end keyframes while better preserving the prompt-specified motion.

As shown in Table~\ref{tab:short_horizon}, our keyframe-conditioned model achieves substantially lower vertex RMSE on unconstrained frames than both the sampling-time replacement baseline and the released open model, demonstrating stronger controllability and more accurate motion in-betweening. These results suggest that effective keyframe conditioning should be learned during training, rather than approximated by simple latent replacement at inference time.

We further evaluate motion quality using VBench. Table~\ref{tab:short_horizon} shows that CompletionAny4D consistently improves the VBench scores over the baseline, while the open model achieves the highest VBench scores in the setting without keyframe constraints. In addition, we conduct a user study comparing our method, the baseline, and AnimateAnyMesh~\cite{Wu_2025_ICCV}, where both our method and the baseline are conditioned only on the start and end keyframes. As shown in Table~\ref{tab:userstudy}, our method is preferred over the open model in text alignment and spatiotemporal controllability, and also outperforms the baseline across the overall perceptual metrics. Taken together, these results demonstrate that our method achieves the strongest spatiotemporal controllability and text alignment among the compared methods, while maintaining competitive motion quality both quantitatively and qualitatively.

We evaluated the inference time and as shown in Table~\ref{tab:vertex_resolution}, our framework achieves faster inference for time compare to the AnimateAnyMesh~\cite{Wu_2025_ICCV}.
\subsubsection{Keyframe conditioning for long horizons.}
In real-world applications, long-horizon motion generation is often more important than short-horizon synthesis. We therefore evaluate whether our 16-frame keyframe-conditioned frame-wise Rectified Flow model can generalize to longer sequences by testing it on 31-frame motions from DyMesh32.

To construct evaluation samples, we concatenate two consecutive 16-frame samples from DyMesh16 and remove the duplicated boundary frame, yielding a continuous 31-frame mesh sequence. All methods are evaluated under a two-stage generation protocol with a one-frame overlap: the first stage generates the first 16 frames, and the second stage generates the last 16 frames including the shared middle frame.

For AnimateAnyMesh~\cite{Wu_2025_ICCV}, the first stage generates the initial 16-frame motion from the initial mesh and the first text prompt. In the second stage, we apply center-scale normalization to the last generated mesh of the first stage, use it as the canonical mesh, and generate the remaining 16 frames conditioned on the second text prompt. For our method and the baseline, we instead use sparse keyframe constraints at frames 1, 16, and 31. The first stage is conditioned on the meshes at frames 1 and 16 together with the first text prompt, and the second stage is conditioned on the meshes at frames 16 and 31 together with the second text prompt.
\begin{figure}
    \centering
    \includegraphics[width=0.99\linewidth]{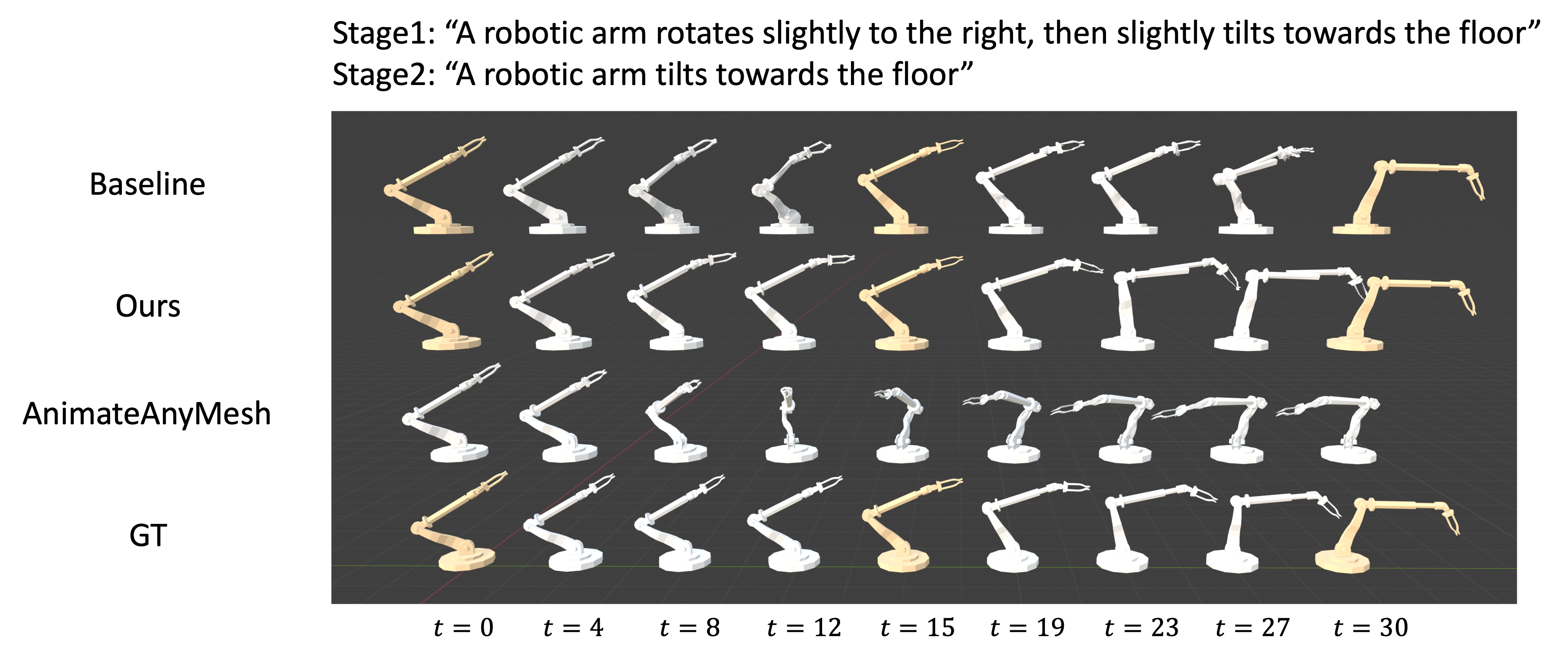}
    \caption{A qualitative result on Dymesh32 dataset with AR generation by AnimateAnyMesh~\cite{Wu_2025_ICCV} and motion in-betweenings. For visualization, we sampled 6 frames from the non-keyframes, the orange colored meshes are used as the keyframes for each methods and stages.}
    \label{fig:main_31}
\end{figure}

Qualitative results are shown in Figure~\ref{fig:main_31}. AnimateAnyMesh tends to accumulate large rotational errors in the first stage, which propagate to the second stage and cause substantial deviation from the ground-truth motion. In contrast, our method preserves better motion consistency across both stages, suggesting that frame-wise keyframe conditioning provides a more stable way to extend generation beyond the training horizon.

We next evaluate motion controllability and generation quality using quantitative metrics. As shown in Table~\ref{tab:long_horizon}, our method substantially improves vertex RMSE, Chamfer distance, and Dynamic Time Warping (DTW), indicating stronger geometric accuracy and temporal alignment under sparse keyframe control. Although AnimateAnyMesh remains strong on the VBench metrics, our method achieves competitive scores and obtains a higher Aesthetic Quality score in the long-horizon setting.

We further conduct a user study to assess perceptual motion quality. As shown in Table~\ref{tab:userstudy}, CompletionAny4D achieves the best overall perceptual performance for the long-horizon settings, indicating that users find our results more natural, shape-consistent, text-aligned, and controllable in space and time. Overall, these results suggest that our method can extend sparse-keyframe-guided 4D mesh generation to longer horizons, even though it is trained only on short sequences.

We evaluated the inference time and as shown in Table~\ref{tab:vertex_resolution}, our framework achieves faster inference time compare to the AnimateAnyMesh~\cite{Wu_2025_ICCV}.

\subsubsection{Frame-wise mesh VAE.}
We compare our frame-wise mesh VAE with DyMeshVAE~\cite{Wu_2025_ICCV} on the same curated DyMesh16 subset under matched preprocessing and training settings, and additionally report the released DyMeshVAE as a reference. 

As shown in Table~\ref{tab:framewiseVAE}, the frame-wise formulation introduces some reconstruction trade-off in RMSE and Chamfer distance relative to the sequence-wise DyMeshVAE. 

We consider this trade-off expected, since preserving frame-aligned latents is necessary for direct keyframe clamping and masked keyframe conditioning. Importantly, the DyMeshVAE retrained on our subset performs comparably to the released model, supporting the adequacy of the curated subset for reliable VAE training. 
Together with the downstream gains in controllability, these results indicate that the frame-wise VAE is the more suitable design for sparse keyframe-conditioned 4D mesh generation.
\begin{table}[t]
\centering
\caption{VAE experiments on our curated DyMesh16 dataset selected shape-agnostic 100 samples.}
\label{tab:framewiseVAE}

{\scriptsize
\setlength{\tabcolsep}{3pt}        
\renewcommand{\arraystretch}{0.95} 

\begin{tabular*}{\columnwidth}{@{\extracolsep{\fill}}lccc@{}}
\toprule
Method & Open & RMSE $\downarrow$ & Chamfer $\downarrow$ \\
\midrule
DyMeshVAE~\cite{Wu_2025_ICCV} & \xmark & \textbf{0.003384} & \textbf{0.006717} \\
\midrule
DyMeshVAE~\cite{Wu_2025_ICCV} (Trained) & \cmark & \underline{0.003542} & \underline{0.007481} \\
\textbf{Ours} & \cmark & 0.004652 & 0.008463 \\
\bottomrule
\end{tabular*}
}
\end{table}
\section{Limitations} CompletionAny4D has three main limitations. First, due to its latent rectified flow design, our framework cannot directly control partial mesh regions or selected point trajectories across the generated frames. Consequently, it is also unable to control intermediate meshes between keyframes. A potential way to address this issue is to perform decoder-based zero-shot test-time optimization. 

Second, the use of a shared shape-structure latent across all generated frames limits the degrees of freedom available for modeling topological changes. However, topology changes frequently occur in real-world physical dynamics. Addressing this limitation would require a neural network that can encode shape features, structural information, and possible topology variations while preserving frame-aligned representations. 

Third, our rectified flow model can generate only fixed-length sequences in a single pass. Therefore, it cannot fully model long sequences over the entire temporal span from start to end. One possible extension is to incorporate previously generated motions and future keyframes into the rectified flow model through an additional neural network or lightweight adapters.

\section{Conclusion}
We introduced CompletionAny4D, the first feed-forward framework for spatiotemporally controllable 4D mesh motion in-betweening over arbitrary meshes. Extensive experiments show that CompletionAny4D enables effective motion control and achieves strong performance on both short- and long-horizon generation, despite being trained only on short-horizon data. These results suggest that explicit spatiotemporal conditioning is a promising direction for controllable 4D mesh animation, when text prompts are insufficient to specify complex motions.

\bibliographystyle{splncs04}

\clearpage
\appendix
\renewcommand{\theHsection}{supp.\Alph{section}}
\renewcommand{\theHsubsection}{supp.\Alph{section}.\arabic{subsection}}
\renewcommand{\theHsubsubsection}{supp.\Alph{section}.\arabic{subsection}.\arabic{subsubsection}}
\def\COMPLETIONANYFOURDMAIN{}
\definecolor{oracleRed}{RGB}{220, 50, 47} 

\newcolumntype{P}[1]{>{\RaggedRight\arraybackslash}p{#1}}
\newcommand{\InBText}{\textcolor{eccvblue}{\textit{InB}}}
\newcommand{\OracleText}{\textcolor{oracleRed}{\textit{(Oracle)}}}
\newcommand{\InBOracle}[2]{\textcolor{eccvblue}{#1}\,\textcolor{oracleRed}{(#2)}}
\newcommand{\NA}{\textemdash}
\definecolor{myblue}{RGB}{232,242,252}
\newcommand{\BlueRow}{%
  \rowcolor{myblue}[2.5pt][2.5pt]%
  \rule[-0.55ex]{0pt}{2.2ex}%
}
\definecolor{oracleRed}{RGB}{220, 50, 47} 
\definecolor{eccvblue}{RGB}{0,77,153} 
\setcounter{table}{0}
\renewcommand{\thetable}{S\arabic{table}}
\section{Comparision with relevant SoTA 4D generative models}
We compared our method with several relevant 4D generation methods. As shown in ~\ref{tab:prior_quantitative_comparison}, our method achieves SoTA results in quantitative and qualitative.

\begin{table*}[t]
\centering
\caption{
Comparison with relevant SoTA 4D-Gen methods.
(N/A: Undefined. \InBText{}: FramePack~\cite{zhang2026frame}, \OracleText{}: GT video input.)
}
\label{tab:prior_quantitative_comparison}

\tiny
\setlength{\tabcolsep}{0.6pt}
\renewcommand{\arraystretch}{0.84}
\resizebox{\textwidth}{!}{%
\begin{tabular}{@{}l*{13}{c}@{}}
\toprule
\textbf{Method / setting}
& \textbf{RMSE$\downarrow$}
& \textbf{CD$\downarrow$}
& \textbf{DTW$\downarrow$}
& \multicolumn{3}{c}{\textbf{VBench$\uparrow$}}
& \textbf{Speed$\downarrow$}
& \textbf{PGB (GB)$\downarrow$}
& \multicolumn{5}{c}{\textbf{User study}} \\
\cmidrule(lr){5-7}\cmidrule(lr){10-14}
&
&
&
& \textbf{I2V$\uparrow$}
& \textbf{M.sm$\uparrow$}
& \textbf{AestQ.$\uparrow$}
&
&
& \textbf{Natur$\uparrow$}
& \textbf{Text$\uparrow$}
& \textbf{Shape$\uparrow$}
& \textbf{Control$\uparrow$}
& \textbf{Div$\uparrow$} \\
\midrule

\multicolumn{14}{@{}l}{
    \textit{Video-to-4D / } \InBText{} \OracleText{}
} \\
DreamGaussian4D (arXiv 2023)
& N/A
& N/A
& N/A
& \InBOracle{0.90}{0.91}
& \InBOracle{0.99}{0.99}
& \InBOracle{0.42}{0.41}
& 18 min
& 21
& \InBOracle{2.24}{2.14}
& \InBOracle{2.89}{2.74}
& \InBOracle{2.56}{2.41}
& \InBOracle{2.34}{2.28}
& N/A \\

4DGen (arXiv 2023)
& N/A
& N/A
& N/A
& \InBOracle{\textbf{0.96}}{\textbf{0.96}}
& \InBOracle{0.99}{0.99}
& \InBOracle{0.42}{0.41}
& 1.2 hours
& 14
& \InBOracle{2.67}{2.50}
& \InBOracle{1.82}{2.21}
& \InBOracle{\underline{3.29}}{3.19}
& \InBOracle{1.76}{2.08}
& N/A \\

SC4D (ECCV 2024)
& N/A
& N/A
& N/A
& \InBOracle{0.91}{0.92}
& \InBOracle{0.99}{0.99}
& \InBOracle{0.41}{0.41}
& 21 min
& 6.0
& \InBOracle{2.42}{2.41}
& \InBOracle{3.04}{2.89}
& \InBOracle{2.73}{2.74}
& \InBOracle{2.10}{2.49}
& N/A \\

GVFD (ICCV 2025)
& N/A
& N/A
& N/A
& \InBOracle{0.89}{0.92}
& \InBOracle{0.97}{0.99}
& \InBOracle{0.36}{0.39}
& 3.9 min
& 14
& \InBOracle{1.93}{1.88}
& \InBOracle{1.86}{2.09}
& \InBOracle{2.31}{2.41}
& \InBOracle{1.44}{1.59}
& N/A \\

SV4D 2.0 (ICCV 2025)
& N/A
& N/A
& N/A
& \InBOracle{0.91}{\underline{0.93}}
& \InBOracle{0.99}{1.0}
& \InBOracle{0.35}{0.34}
& 2.3 min
& 27
& \InBOracle{2.76}{2.30}
& \InBOracle{\underline{3.65}}{3.31}
& \InBOracle{3.24}{2.59}
& \InBOracle{2.54}{2.83}
& N/A \\

\rowcolor{myblue}
ActionMesh (CVPR 2026)
& \InBOracle{0.27}{0.27}
& \InBOracle{0.43}{0.46}
& \InBOracle{0.35}{0.39}
& \InBOracle{0.89}{0.91}
& \InBOracle{1.0}{1.0}
& \InBOracle{0.43}{0.44}
& 3.2 min
& 11
& \InBOracle{2.52}{\underline{3.02}}
& \InBOracle{2.89}{3.17}
& \InBOracle{2.90}{\underline{3.29}}
& \InBOracle{2.63}{\underline{2.99}}
& N/A \\

\midrule

\multicolumn{14}{@{}l}{\textit{3D + Video-to-4D / } \InBText{} \OracleText{}} \\
\rowcolor{myblue}
ActionMesh (CVPR 2026)
& \InBOracle{0.31}{0.33}
& \InBOracle{0.39}{0.41}
& \InBOracle{0.28}{0.30}
& \InBOracle{0.92}{0.91}
& \InBOracle{1.0}{0.99}
& \InBOracle{\underline{0.46}}{\textbf{0.47}}
& 3.2 min
& 11
& \InBOracle{2.90}{2.14}
& \InBOracle{2.56}{1.71}
& \InBOracle{\underline{3.29}}{2.63}
& \InBOracle{2.73}{2.08}
& N/A \\

\rowcolor{myblue}
Motion 3-to-4 (CVPR 2026)
& \InBOracle{0.31}{0.27}
& \InBOracle{0.50}{0.46}
& \InBOracle{0.43}{0.43}
& \InBOracle{0.89}{0.91}
& \InBOracle{1.0}{1.0}
& \InBOracle{0.42}{0.43}
& \underline{8 sec}
& \textbf{4.3}
& \InBOracle{2.09}{2.77}
& \InBOracle{2.34}{2.82}
& \InBOracle{2.32}{3.00}
& \InBOracle{2.15}{2.84}
& N/A \\

\midrule

TC4D (ECCV 2024)
& N/A
& N/A
& N/A
& 0.83
& 0.99
& \underline{0.46}
& 31 hours
& 23
& 2.03
& 1.74
& \textbf{3.67}
& 1.05
& N/A \\

\rowcolor{myblue}
AnyMoLe (CVPR 2025)
& \underline{0.26}
& \underline{0.33}
& \underline{0.17}
& 0.90
& 0.99
& 0.45
& 2.5 hours
& 45
& 2.40
& 2.26
& 2.94
& 2.58
& 1.50 \\

\midrule

FramePack (NeurIPS 2025)
& N/A
& N/A
& N/A
& N/A
& N/A
& N/A
& 6.4 min
& 30
& N/A
& N/A
& N/A
& N/A
& \textbf{3.70} \\

\midrule
\rowcolor{myblue}
\textbf{Ours}
& \textbf{0.19}
& \textbf{0.25}
& \textbf{0.11}
& 0.90
& 0.99
& 0.44
& \textbf{4.0 sec}
& \underline{7.5}
& \textbf{3.10}
& \textbf{4.11}
& 3.22
& \textbf{3.83}
& \underline{3.10} \\

\bottomrule
\end{tabular}%
}
\end{table*}

\section{Protocol Clarifications for the Main-Paper Setting}

\subsection{How to read the experimental setting}

\subsubsection{Main-paper setting and purpose of this supplementary}
As defined in Sec.~3.1 of the main paper, our task is text-and-mesh-driven motion in-betweening with a mandatory reference mesh $M_0$ and optional sparse keyframes.
Each sequence has fixed topology, while topology may vary across sequences and categories.
This supplementary does not redefine that task setting; instead, it explains how that already-defined setting determines the detailed evaluation protocol, the scope of the primary baselines, and the interpretation of the longer-rollout experiment.

\begin{table}[!htbp]
\caption{
Related Work summaries, emphasizing whether each method is a motion in-betweening method and whether it is feed-forward.
$^\ast$Feed-forward denotes amortized test-time inference without per-scene optimization or scene-specific training.
}
\centering
\small
\setlength{\tabcolsep}{8pt}
\renewcommand{\arraystretch}{1.10}
\begin{tabular}{l c c c}
\toprule
Method &
\shortstack{In-betweening} &
\shortstack{Feed-forward$^\ast$} &
Venue \\
\midrule

\multicolumn{4}{l}{{Motion in-betweening methods}} \\

AnyMoLe~\cite{yun2025anymole}
& \cmark & \xmark & CVPR'25 \\

In-2-4D~\cite{Nag2025In2_4D}
& \cmark & \xmark & SA'25 \\

\textbf{Ours}
& \cmark & \cmark & -- \\

\midrule
\multicolumn{4}{l}{{Other related 4D generation / control methods}} \\

SC4D~\cite{Wu2024SC4D}
& \xmark & \xmark & ECCV'24 \\

DreamMesh4D~\cite{Li2024DreamMesh4D}
& \xmark & \xmark & NeurIPS'24 \\

TC4D~\cite{Bahmani2024TC4D}
& \xmark & \xmark & ECCV'24 \\

\bottomrule
\end{tabular}
\label{tab:rw_inbetweening_ff}
\end{table}
\subsubsection{Core methodological claim under this setting}
The main methodological claim of this paper is not overall superiority over all 4D generators.
Rather, under the main-paper setting above, the central claim is that frame-aligned latents plus training-time masked keyframe conditioning are necessary for controllable sparse-keyframe 4D motion in-betweening.

\subsubsection{How to interpret the main comparisons}
AnimateAnyMesh is included as a strong released reference rather than a same-split retrained baseline.
Our controlled comparison for isolating the effect of training-time keyframe conditioning is \textbf{Baseline vs. Ours} under the same open split.
The released AnimateAnyMesh model serves as a closed-protocol reference with an undisclosed split.
Accordingly, comparisons to AnimateAnyMesh should be read as reference-score comparisons rather than same-split controlled ablations.

\subsubsection{Scope of primary baselines}
We focus on feed-forward text-and-mesh-conditioned dynamic mesh generation.
We therefore use as primary baselines methods that share this inference regime and conditioning type, while methods requiring image/video inputs or per-instance test-time optimization are discussed as related work but are not task-matched primary baselines.

\subsubsection{Why the quantitative metrics exclude conditioned keyframes}
\noindent\textbf{Keyframe preservation.}
For quantitative evaluation, conditioned keyframes are explicitly replaced with the ground-truth keyframes in the final assembled sequence.
Accordingly, all motion-accuracy metrics reported in the main paper are computed only on unconstrained non-keyframes.
This protocol evaluates the accuracy of the generated in-betweening frames rather than trivially fixed inputs.

\subsubsection{How to read the DyMesh32 experiment}
\noindent\textbf{Interpretation of the DyMesh32 experiment.}
We do not claim single-pass modeling of arbitrarily long sequences.
Our DyMesh32 experiment instead evaluates a practical beyond-training-horizon extension: a two-stage 31-frame rollout with sparse keyframes at frames 1, 16, and 31 in the paper's one-based indexing (equivalently 0, 15, and 30 in zero-based indexing used by the implementation).
The claim supported by this experiment is that sparse keyframe conditioning stabilizes stage-to-stage rollout and reduces drift beyond the 16-frame training horizon.

\section{Evaluation Protocols}

\subsection{Metrics and rendering}

\subsubsection{Quantitative Metrics.}
We follow the implementation used in our evaluation code.
Let $\mathrm{pred}$ and $\mathrm{gt}$ denote the predicted and ground-truth mesh sequences.
Let $\mathrm{pred}_{\mathrm{use}}$ and $\mathrm{gt}_{\mathrm{use}}$ denote the subsets selected for evaluation (e.g., all frames, keyframes only, or non-keyframes only).
Unless otherwise noted, the metrics reported in the main paper are computed on non-keyframes only.

\subsubsection{RMSE.}
We first compute the element-wise RMSE
\begin{equation}
\mathrm{RMSE}=\sqrt{{(\mathrm{pred}-\mathrm{gt})}^2}.
\end{equation}
\subsubsection{Chamfer Distance.}
For each frame, let $A_i=\mathrm{pred}[i]$ and $B_i=\mathrm{gt}[i]$ denote the vertex point sets, and let $T=\min(T_{\mathrm{pred}},T_{\mathrm{gt}})$. Using the unsquared Euclidean distance, the sequence-level score is
\begin{equation}
\mathrm{CD}_{\mathrm{mean\_aligned}}
=
\frac{1}{T}\sum_{i=0}^{T-1}
\left(
\frac{1}{|B_i|}\sum_{b\in B_i}\min_{a\in A_i}\|a-b\|_2
+
\frac{1}{|A_i|}\sum_{a\in A_i}\min_{b\in B_i}\|a-b\|_2
\right).
\end{equation}

\subsubsection{Dynamic Time Warping (DTW).}
Let $\Pi_{n,m}$ be the set of monotone warping paths from $(0,0)$ to $(n-1,m-1)$, where $n=T_{\mathrm{pred}}$ and $m=T_{\mathrm{gt}}$. Our path-length-normalized DTW on centroid trajectories is
\begin{equation}
\begin{aligned}
\mathrm{DTW}_{\mathrm{centroid\_norm}}
&=
\frac{1}{|\pi^\star|}
\sum_{(i,j)\in\pi^\star}\|\mathbf{c}^{p}_{i}-\mathbf{c}^{g}_{j}\|_2,\quad
\pi^\star
=
\arg\min_{\pi\in\Pi_{n,m}}
\sum_{(i,j)\in\pi}\|\mathbf{c}^{p}_{i}-\mathbf{c}^{g}_{j}\|_2,\\
\mathbf{c}^{p}_{i}
&=
\frac{1}{V}\sum_{v=1}^{V}\mathrm{pred}[i,v,:],\qquad
\mathbf{c}^{g}_{j}
=
\frac{1}{V}\sum_{v=1}^{V}\mathrm{gt}[j,v,:].
\end{aligned}
\end{equation}

\subsubsection{VBench Rendering Protocol}
Because the dataset does not provide textures, we assign a single light-purple material to all meshes and render videos from six fixed orthographic views.
Specifically, we use the Blender EEVEE backend through the \texttt{bpy} interface (dependency: \texttt{bpy==5.0.0}) with the following settings:
\begin{itemize}
    \item Resolution: $512\times512$
    \item Frame rate: 10 FPS
    \item Camera: six fixed views looking at the origin, with $(\mathrm{azimuth},\mathrm{elevation})\in\{(0,0),(90,0),(180,0),(270,0),(0,90),(0,-90)\}$, corresponding to front, right, back, left, top, and bottom
    \item Camera model: orthographic camera with \texttt{ortho\_scale}=2.5
    \item Mesh centering: none at render time; the FBX coordinates are rendered as-is
    \item Mesh scaling: none at render time; apparent scale is controlled by the fixed orthographic camera scale
    \item Lighting: nine SUN lights (eight surrounding lights plus one top light), with shadows enabled
    \item Background: solid dark gray, RGBA $=(0.05,0.05,0.05,1.0)$, strength $=1.0$
    \item Material: texture-free Principled BSDF
    \item Mesh color: $(0.55,0.25,0.78,1.00)$
\end{itemize}
For VBench evaluation, each generated mesh sequence is rendered into six videos using the fixed views above.
We compute each VBench sub-metric independently on the six rendered videos and report the arithmetic mean across views.

\begin{figure}[t]
    \centering
    \setlength{\tabcolsep}{2pt}
    \begin{tabular}{ccc}
        \includegraphics[width=0.31\linewidth]{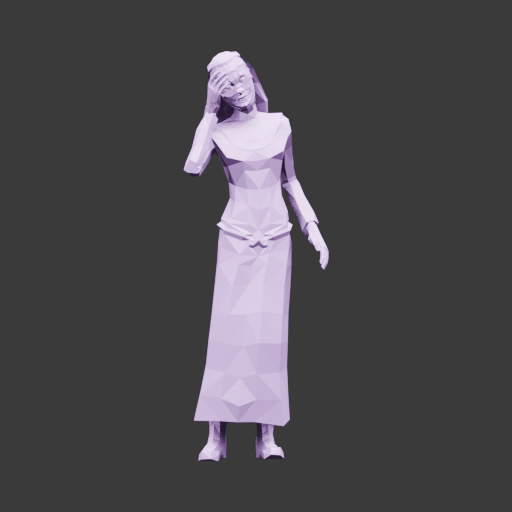} &
        \includegraphics[width=0.31\linewidth]{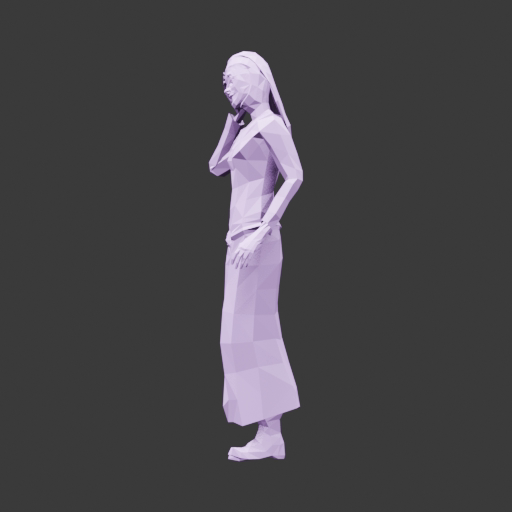} &
        \includegraphics[width=0.31\linewidth]{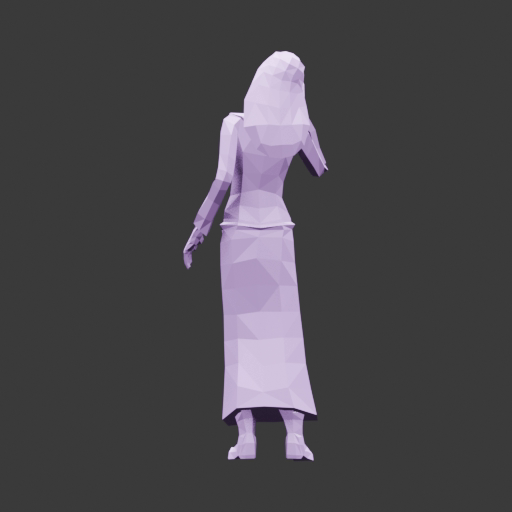} \\
        {\small Front} & {\small Right side} & {\small Back} \\
        \includegraphics[width=0.31\linewidth]{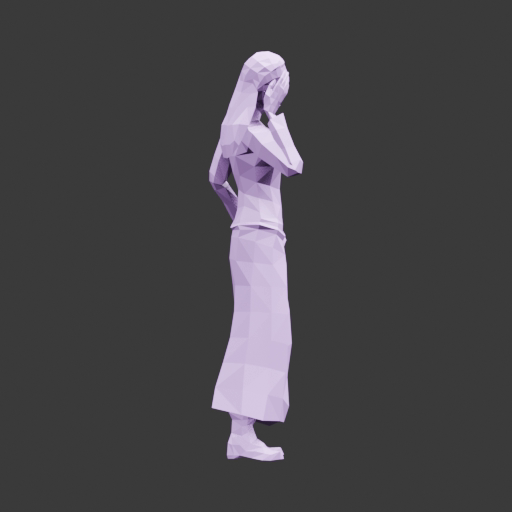} &
        \includegraphics[width=0.31\linewidth]{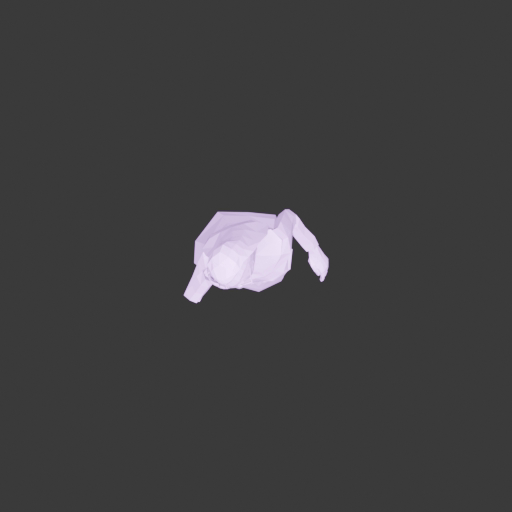} &
        \includegraphics[width=0.31\linewidth]{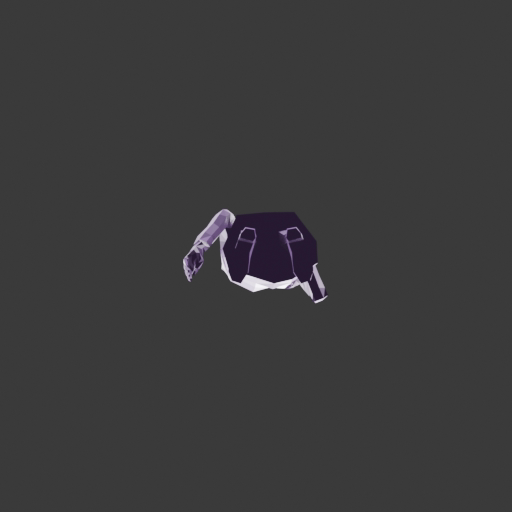} \\
        {\small Left side} & {\small Top} & {\small Bottom}
    \end{tabular}
    \caption{Representative frames from the six fixed orthographic views used for VBench evaluation: front, right side, back, left side, top, and bottom. All methods are rendered with the same camera, lighting, background, material, and frame rate settings. VBench is computed on the six rendered videos from these viewpoints, and the reported score is the mean across views.}
    \label{fig:vbench_render_examples}
\end{figure}

\subsection{Human evaluation}

\subsubsection{User Study Protocol and Statistical Tests}
We follow the user-study setup described in the main paper and evaluate four criteria: Natur, Text, Shape, and Control.
At the time of writing this draft, the detailed participant counts, exact example counts, presentation interface snapshots, and statistical test results are being consolidated into the final supplementary package.
The intended reporting protocol is as follows:
\begin{itemize}
    \item 5-point Likert ratings for the four criteria above
    \item randomized presentation order across methods
    \item blinded evaluation without method names
    \item paired non-parametric analysis using a Friedman test followed by post-hoc Wilcoxon signed-rank tests with Holm correction when appropriate
\end{itemize}

\begin{figure}[t]
    \centering
    \includegraphics[width=0.99\linewidth]{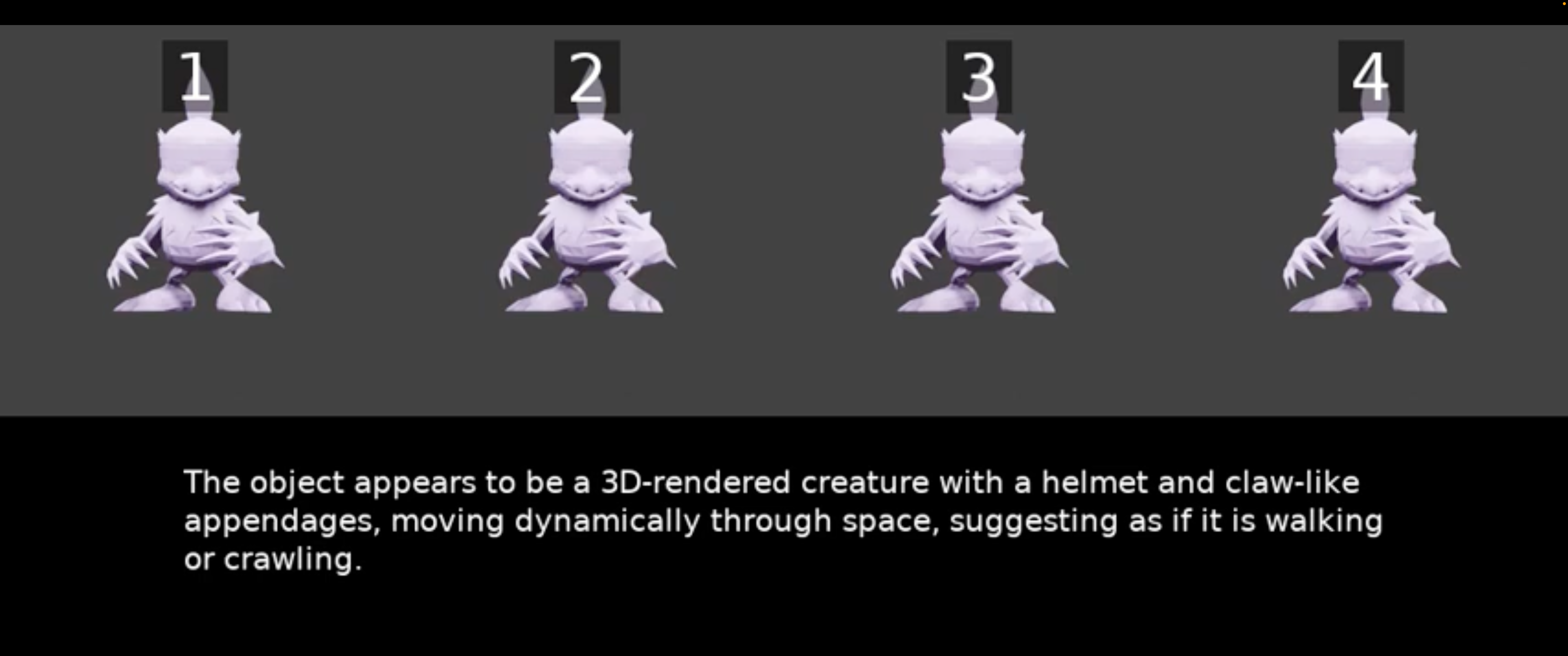}
    \caption{Representative snapshots used in the user study. This figure should match the actual evaluation interface and examples.}
    \label{fig:user_study_snapshots}
\end{figure}

\FloatBarrier
\section{Training and Implementation Details}

\subsection{Data and model variants}

\subsubsection{Dataset Curation and Split}
We train on the DyMesh16 subset used in the main paper.
We restrict the vertex count to 512--4096, resulting in approximately 260K samples in total with an 80/20 train/validation split.
The main paper already reports this curation protocol; here we provide the implementation-level hyperparameters for each model variant.

\subsubsection{Model Variants}
We use four main model variants in our experiments:
(i) a keyframe-conditioned RF model (our method),
(ii) a non-keyframe RF model (baseline without training-time keyframe conditioning),
(iii) a frame-wise VAE, and
(iv) a sequence-wise VAE.

\subsection{Training configurations}
Tables~\ref{tab:rf_config} and~\ref{tab:vae_config} summarize the implementation details of the four model variants used in this supplementary.
Table~\ref{tab:rf_config} reports the RF configurations for our keyframe-conditioned model and the non-keyframe baseline, organized into optimization, inference, and architecture groups.
Table~\ref{tab:vae_config} reports the corresponding optimization, runtime, and architecture settings for the frame-wise and sequence-wise VAE variants, including their latent tensor shapes.

\subsubsection{Rectified Flow Models}
Table~\ref{tab:rf_config} lists the optimization hyperparameters, ODE-solver settings at inference, and architectural details of the two RF variants.
In particular, the rows under {Optimization} specify the training hyperparameters, the rows under {Inference} specify the solver and sampling protocol, and the rows under {Architecture} specify the backbone and latent input/output shapes.
\begin{table*}[!t]
\centering
\caption{Compact summary of RF model configurations. The rows are grouped by role so that the table can be read as a concise procedural summary of each model variant.}
\label{tab:rf_config}
\footnotesize
\setlength{\tabcolsep}{4pt}
\renewcommand{\arraystretch}{1.08}
\begin{tabularx}{\textwidth}{P{0.23\textwidth}XX}
\toprule
\textbf{Item} & \textbf{Keyframe-conditioned RF (ours)} & \textbf{Non-keyframe RF (baseline)} \\
\midrule
\tablegroup{Optimization}
Batch size & 256 & 256 \\
Optimizer & AdamW, betas $(0.9,0.99)$, weight decay $0.01$ & AdamW, betas $(0.9,0.99)$, weight decay $0.01$ \\
Learning rate & Constant $2\times10^{-4}$, no warmup / no schedule & Constant $2\times10^{-4}$, no warmup / no schedule \\
Gradient clipping & 1.0 & 1.0 \\
Precision & FP32 (inferred) & FP32 (inferred) \\
Seed & Training seed not explicitly logged; sampler seed $=42$, validation render seed $=666$ & Training seed not explicitly logged; sampler seed $=42$, validation render seed $=666$ \\
\tablegroup{Inference}
ODE solver & \code{torchdiffeq.odeint(method='midpoint')} & \code{torchdiffeq.odeint(method='midpoint')} \\
Inference steps & 64 & 64 \\
CFG scale & 1.0 in validation rendering & 1.0 in validation rendering \\
\tablegroup{Architecture}
Backbone & MMDiT & MMDiT \\
\# blocks & 12 & 12 \\
Hidden dim. & 512 & 512 \\
\# heads & 8 & 8 \\
Positional encoding & Timestep embedding (sinusoidal PE) & Timestep embedding (sinusoidal PE) \\
Text encoder & CLIP ViT-L/14 & CLIP ViT-L/14 \\
Latent I/O shape & Output: $16\times12=192$; input: $221=(T+1)\times(C+1)$ & Output: $16\times12=192$; input: $204=(T+1)\times C$ \\
\bottomrule
\end{tabularx}
\end{table*}

\subsubsection{VAE Models}
Table~\ref{tab:vae_config} summarizes the two VAE variants.
The rows under {Optimization} report the training hyperparameters and loss, and the rows under {Architecture} report the backbone, hidden dimension, positional encoding, and latent tensor shape.
\begin{table*}[!t]
\centering
\caption{Compact summary of VAE configurations.}
\label{tab:vae_config}
\footnotesize
\setlength{\tabcolsep}{4pt}
\renewcommand{\arraystretch}{1.08}
\begin{tabularx}{\textwidth}{P{0.23\textwidth}XX}
\toprule
\textbf{Item} & \textbf{Frame-wise VAE} & \textbf{Sequence-wise VAE} \\
\midrule
\tablegroup{Optimization}
Batch size & 256 & 128 \\
Optimizer & AdamW, default betas $(0.9,0.999)$, weight decay $10^{-4}$ & AdamW, default betas $(0.9,0.999)$, weight decay $10^{-4}$ \\
Learning rate & Constant $10^{-4}$, no warmup / no schedule & Constant $10^{-4}$, no warmup / no schedule \\
Gradient clipping & 1.0 & 1.0 \\
Precision & FP32 (\code{amp=false}) & FP32 (\code{amp=false}) \\
Seed & 666 & 666 \\
Loss & $\mathcal{L}=\mathcal{L}_{\mathrm{recon}} + 10^{-4}\mathcal{L}_{\mathrm{KL}}$ & $\mathcal{L}=\mathcal{L}_{\mathrm{recon}} + 10^{-4}\mathcal{L}_{\mathrm{KL}}$ \\
\tablegroup{Architecture}
Backbone & DyMeshVAE & DyMeshVAE \\
Hidden dim. & 128 & 256 \\
\# heads & 8 & 8 \\
Positional encoding & PointEmbed + TrajEncoding (sin/cos Fourier) & PointEmbed + TrajEncoding (sin/cos Fourier) \\
Latent tensor shape & $x=\mathrm{concat}(x_0,x_t)$, latent dim.\ $12+12=24$, shape approx.\ $[B,512,24]$ & $x_0$ and $x_t$ each use latent dim.\ 32, total 64 channels, shape approx.\ $[B,512,64]$ \\
\bottomrule
\end{tabularx}
\end{table*}

\FloatBarrier
\section{Long-Horizon Extension Details}

\subsubsection{Two-Stage 31-Frame Rollout}
All methods are evaluated under a two-stage generation protocol with a one-frame overlap.
In the paper's one-based indexing, stage~1 covers frames 1--16 and stage~2 covers frames 16--31; in the implementation, this corresponds to zero-based indexing with overlap at global frame~15.
Although the benchmark name is DyMesh32, the stitched output contains 31 distinct frames because the two 16-frame windows share one overlapping frame.
The stage-wise normalization, de-normalization, and final stitching procedure are visualized for the autoregressive rollout baseline in Fig.~\ref{fig:long_horizon_protocol_animate} and for CompletionAny4D in Fig.~\ref{fig:long_horizon_protocol_ours}.

\subsubsection{Two-Stage 31-Frame Rollout}
All methods are evaluated under a two-stage generation protocol with a one-frame overlap.
In the paper's one-based indexing, stage~1 covers frames 1--16 and stage~2 covers frames 16--31.
In the implementation, this corresponds to zero-based indexing with overlap at global frame~15.

\subsubsection{Procedure Summary}
For readability, the two-stage protocol is summarized below.
\begin{enumerate}
    \item Generate stage~1 using the stage-1 prompt and the first set of keyframes.
    \item Let $V^{(1)}_{\mathrm{ov}}$ denote the overlapped stage-1 output frame and compute the stage-1 center $\mathbf{c}_1$ and scale $s_1$ from it.
    \item Normalize the stage-2 local input using $\mathbf{c}_1$ and $s_1$.
    \item Replace the first local stage-2 frame with either the generated overlap frame or the ground-truth overlap frame, depending on the chosen protocol.
    \item Generate stage~2 using the stage-2 prompt and the second set of keyframes.
    \item De-normalize the stage-2 outputs with the same $\mathbf{c}_1$ and $s_1$, and stitch the final result by taking frames 1--15 from stage~1 and frames 16--31 from stage~2 in one-based indexing.
\end{enumerate}
\begin{figure}[t]
    \centering
    \includegraphics[width=0.99\linewidth]{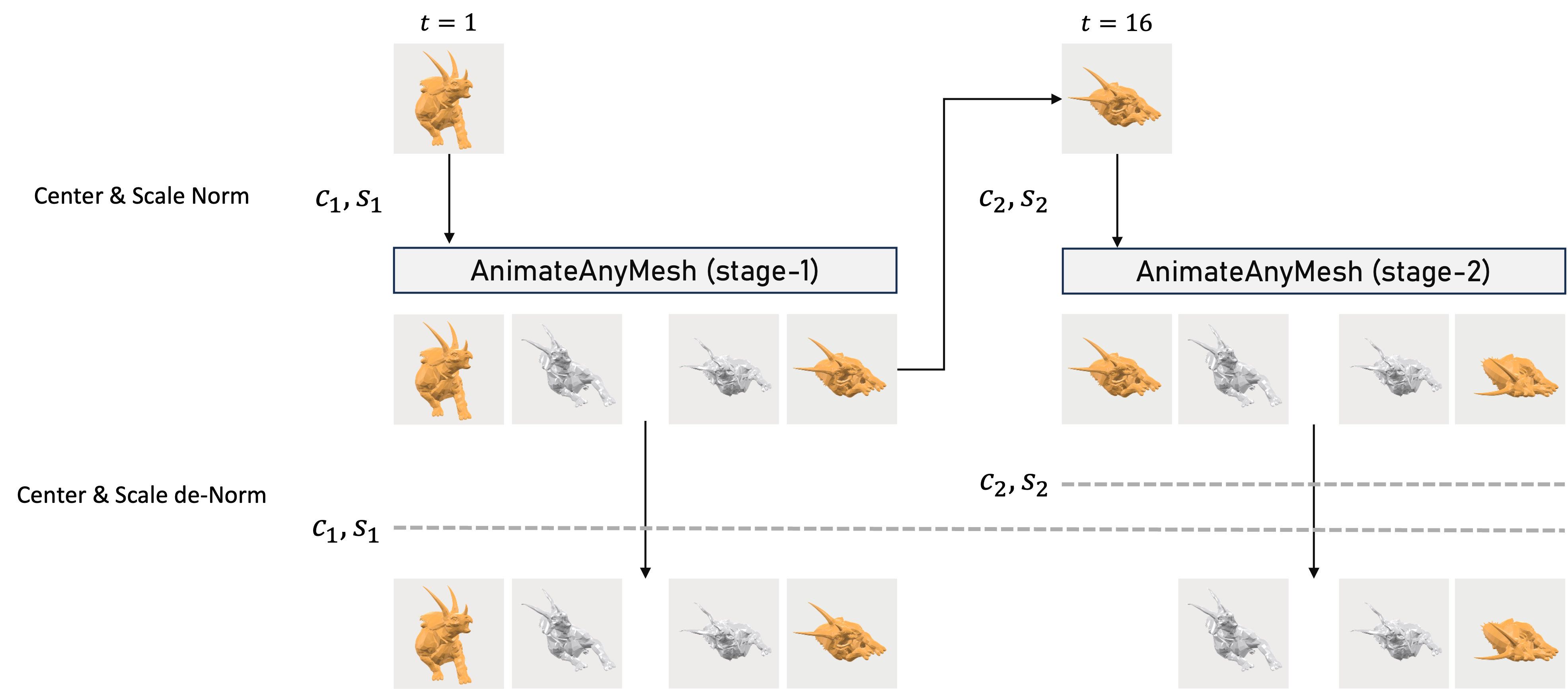}
    \caption{Illustration of the autoregressive 31-frame rollout on the DyMesh32 dataset using AnimateAnyMesh~\cite{Wu_2025_ICCV}. For clarity, text prompts are omitted; the figure focuses on how frames are normalized and denormalized at each stage and how the stage outputs are stitched together. Text prompts are specified independently for each stage.}
    \label{fig:long_horizon_protocol_animate}
\end{figure}

\begin{figure}[t]
    \centering
    \includegraphics[width=0.99\linewidth]{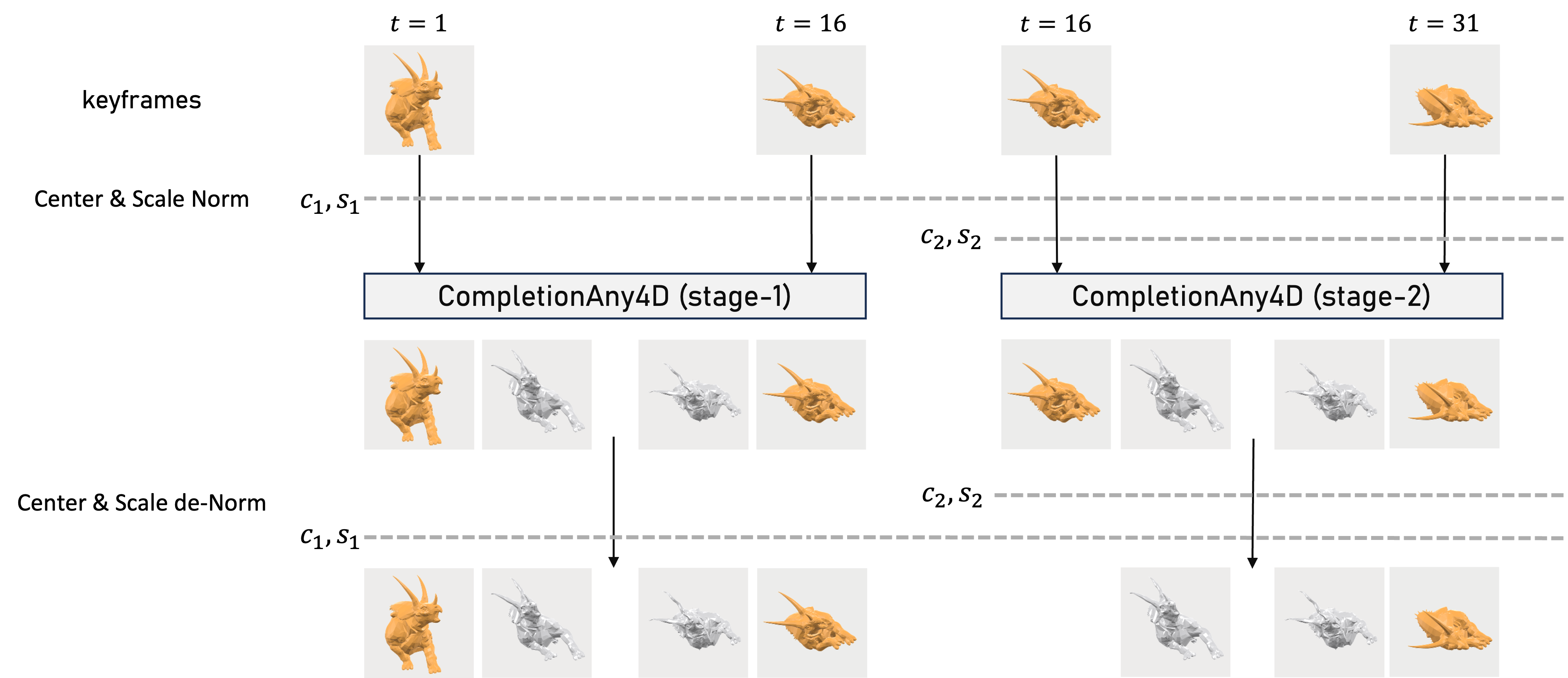}
    \caption{Illustration of the 31-frame longer-horizon generation on the DyMesh32 dataset using CompletionAny4D. For clarity, text prompts are omitted; the figure focuses on how frames are normalized and denormalized at each stage and how the stage outputs are stitched together. Text prompts are specified independently for each stage.}
    \label{fig:long_horizon_protocol_ours}
\end{figure}
\subsubsection{Window Normalization and Stitching}
The implementation performs center-scale normalization when preparing the stage-2 input.
Let $V^{(1)}_{\mathrm{ov}}$ denote the overlapped stage-1 output frame (global frame~15 in zero-based indexing).
We define the stage-1 center and scale as
\begin{equation}
\mathbf{c}_1
=
\frac{\max(V^{(1)}_{\mathrm{ov}})+\min(V^{(1)}_{\mathrm{ov}})}{2},
\qquad
s_1
=
\max\left(\left|V^{(1)}_{\mathrm{ov}}-\mathbf{c}_1\right|\right),
\end{equation}
with a fallback to $1.0$ if $s_1$ is non-positive or non-finite.

Using these stage-1 statistics, the stage-2 local input is normalized as
\begin{equation}
\overline{V}^{(2)}_t
=
\frac{V_{15+t}-\mathbf{c}_1}{s_1},
\qquad t=0,\dots,15,
\end{equation}
where $V_{15+t}$ denotes the global-coordinate mesh for the stage-2 window.

After stage-2 generation in the normalized coordinate system, the outputs are mapped back to the global coordinates by de-normalization:
\begin{equation}
\widehat{V}^{(2)}_{15+t}
=
s_1\,\widehat{\overline{V}}^{(2)}_t+\mathbf{c}_1,
\qquad t=0,\dots,15.
\end{equation}

This normalization/de-normalization flow and the final stage stitching are illustrated in Fig.~\ref{fig:long_horizon_protocol_animate} for the autoregressive rollout baseline and in Fig.~\ref{fig:long_horizon_protocol_ours} for CompletionAny4D.

The final assembled sequence is produced by simple stitching:
\begin{equation}
\mathrm{final}[0{:}15] = \mathrm{stage1}[0{:}15],
\qquad
\mathrm{final}[15{:}31] = \widehat{V}^{(2)}_{15{:}31}.
\end{equation}
Equivalently, in one-based indexing, frames 1--15 are taken from stage~1 and frames 16--31 are taken from the de-normalized stage~2 outputs.

\subsubsection{Prompt Handling and Keyframe Injection}
The stage prompts are chosen as follows:
\begin{itemize}
    \item Stage-1 prompt: \code{--prompt-stage1} if provided; otherwise \code{--prompt}; otherwise \code{source_caption}
    \item Stage-2 prompt: \code{--prompt-stage2} if provided; otherwise \code{--prompt}; otherwise \code{source_caption}
\end{itemize}
When constructing the stage-2 local input, the first frame can be replaced before inference by either the generated overlap frame or the ground-truth overlap frame depending on the selected protocol.

\subsubsection{What This Experiment Supports}
This experiment supports the claim that sparse keyframe conditioning stabilizes stage-to-stage rollout beyond the 16-frame training horizon.
It should not be interpreted as evidence for single-pass modeling of arbitrarily long sequences.

\section{Additional Qualitative Results}

We provide additional qualitative results to further demonstrate the effectiveness of CompletionAny4D.
Specifically, we include five additional short-horizon examples on 16-frame sequences and five additional long-horizon examples on 31-frame sequences.
We further perform inference on approximately 24 long-horizon samples beyond 31 frames.
The corresponding animated results are provided in the supplementary video.

\subsection{Short-Horizon Results (16 Frames)}
We present five additional qualitative examples of short-horizon motion in-betweening on 16-frame sequences.
These examples further demonstrate that our method produces smooth and controllable transitions while preserving the input keyframes and the motion semantics specified by the text prompt.
The corresponding animations are also included in the supplementary video.

\subsection{Long-Horizon Results (31 Frames)}
We present five additional qualitative examples of long-horizon generation on 31-frame sequences.
These results show that our method maintains temporal consistency and spatiotemporal controllability over longer durations under sparse keyframe constraints.
The corresponding animations are also included in the supplementary video.

\subsection{Long-Horizon Results Beyond 31 Frames}
In addition to the 31-frame setting, we perform inference on approximately 24 long-horizon samples beyond 31 frames.
These results further illustrate the behavior of our method on longer temporal horizons.
All of these examples can be viewed in the supplementary video.

\section{Additional Quantitative Results}

\subsection{Statistical Tests for the User Study}

In the main paper, we reported the aggregated 5-point Likert scores from the user study following the evaluation protocol of AnimateAnyMesh.
While these aggregate scores provide a concise summary, they do not reveal the distribution of participant responses or the statistical reliability of the observed differences.
To complement the main-paper results, we therefore present the rating distributions as boxplots and additionally report non-parametric statistical tests.

\begin{figure*}[t]
    \centering
    \includegraphics[width=\textwidth]{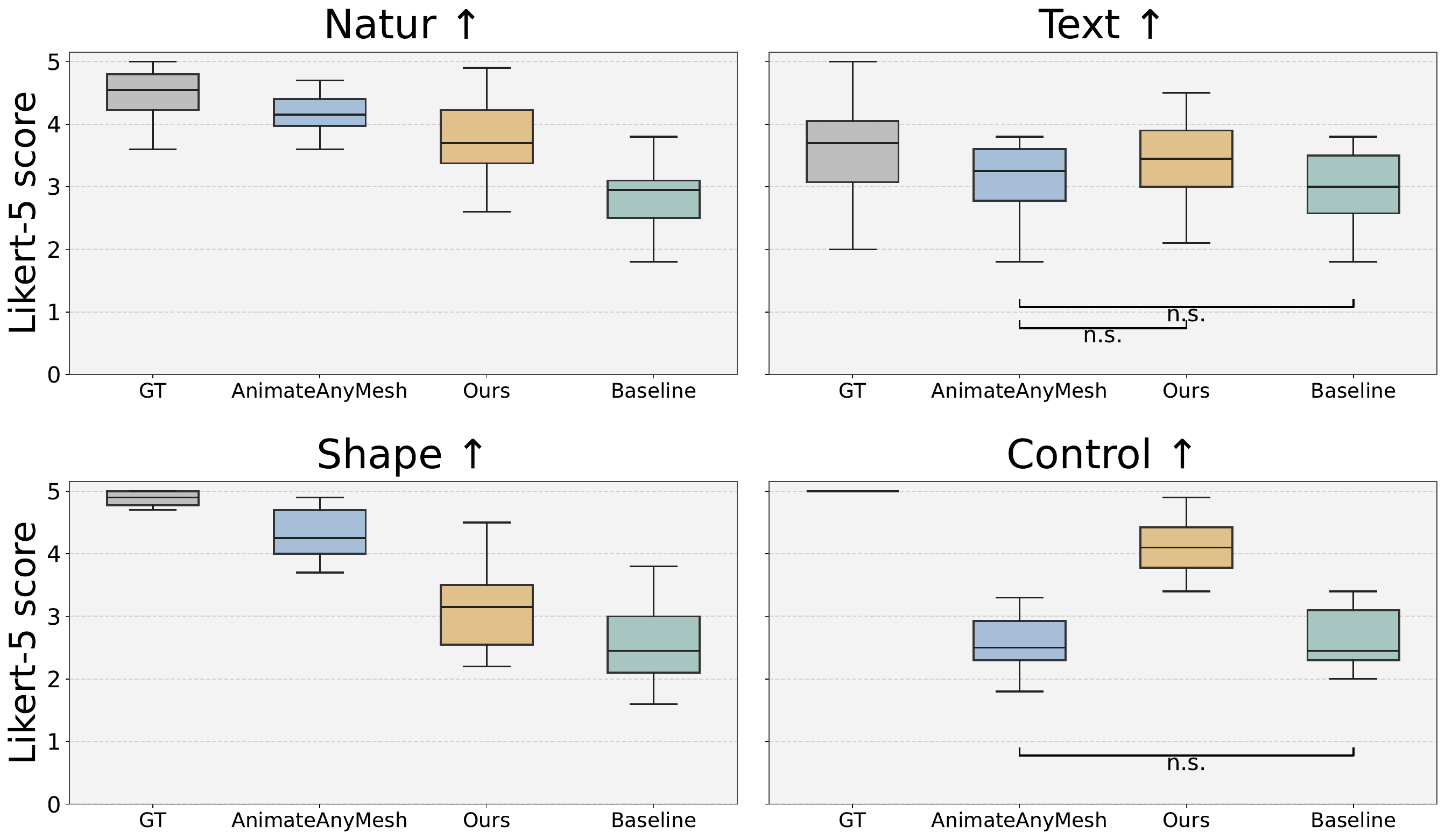}
    \caption{\textbf{5-point Likert score distributions (short-horizon user study).}
    Boxplots summarize participant ratings for \textbf{GT}, \textbf{AnimateAnyMesh}, \textbf{Ours}, and \textbf{Baseline} on \textbf{Natur}$\uparrow$, \textbf{Text}$\uparrow$, \textbf{Shape}$\uparrow$, and \textbf{Control}$\uparrow$.
    Statistical testing uses a Friedman test across the four methods with participant as the blocking factor ($n=20$ participants, $k=4$ methods), followed by paired Wilcoxon signed-rank post-hoc tests with Holm--Bonferroni correction.
    Holm correction is applied across the four questionnaire items and again across the six pairwise comparisons within each item.
    The significance level is $\alpha=0.05$.
    Brackets labeled \emph{n.s.} indicate method pairs that are not significantly different after correction.
    For the short-horizon study, the non-significant pairs are \textbf{AnimateAnyMesh--Baseline} for \textbf{Control}, and \textbf{AnimateAnyMesh--Ours} and \textbf{AnimateAnyMesh--Baseline} for \textbf{Text}.}
    \label{fig:userstudy_boxplots_q2_v9}
\end{figure*}

\begin{figure*}[t]
    \centering
    \includegraphics[width=\textwidth]{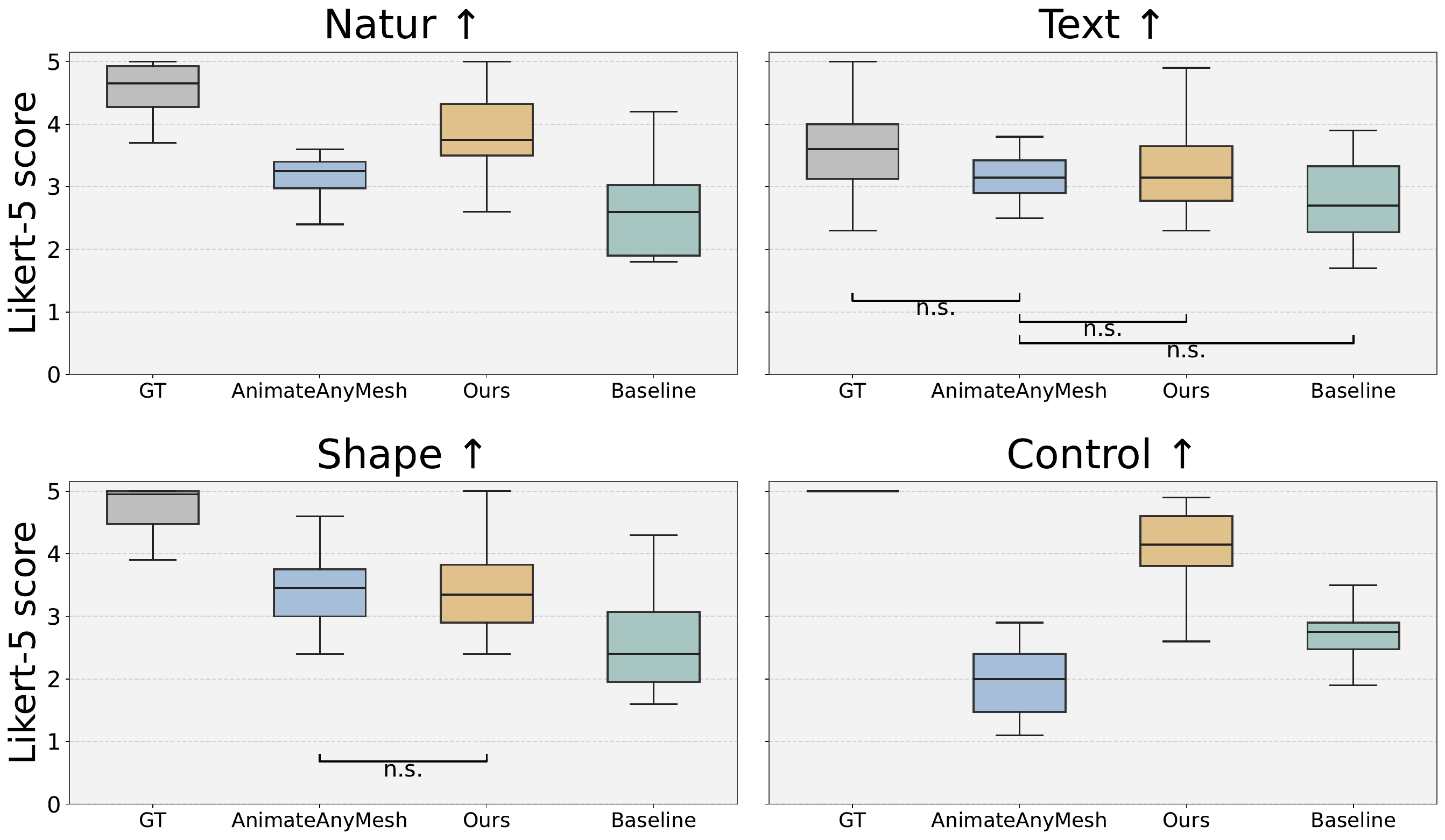}
    \caption{\textbf{5-point Likert score distributions (long-horizon user study).}
    Boxplots summarize participant ratings for \textbf{GT}, \textbf{AnimateAnyMesh}, \textbf{Ours}, and \textbf{Baseline} on \textbf{Natur}$\uparrow$, \textbf{Text}$\uparrow$, \textbf{Shape}$\uparrow$, and \textbf{Control}$\uparrow$.
    Statistical testing uses a Friedman test across the four methods with participant as the blocking factor ($n=20$ participants, $k=4$ methods), followed by paired Wilcoxon signed-rank post-hoc tests with Holm--Bonferroni correction.
    Holm correction is applied across the four questionnaire items and again across the six pairwise comparisons within each item.
    The significance level is $\alpha=0.05$.
    Brackets labeled \emph{n.s.} indicate method pairs that are not significantly different after correction.
    For the long-horizon study, the non-significant pairs are \textbf{AnimateAnyMesh--Ours} for \textbf{Shape}, and \textbf{GT--AnimateAnyMesh}, \textbf{AnimateAnyMesh--Ours}, and \textbf{AnimateAnyMesh--Baseline} for \textbf{Text}.}
    \label{fig:userstudy_boxplots_q1_v9}
\end{figure*}

Figures~\ref{fig:userstudy_boxplots_q2_v9} and~\ref{fig:userstudy_boxplots_q1_v9} show the results of the short-horizon and long-horizon user studies, respectively.
For each study, we compare \textbf{GT}, \textbf{AnimateAnyMesh}, \textbf{Ours}, and \textbf{Baseline} on the four criteria \textbf{Natur}, \textbf{Text}, \textbf{Shape}, and \textbf{Control}.
We first apply a Friedman test across the four methods with participant as the blocking factor.
When the omnibus test is significant, we further conduct paired Wilcoxon signed-rank post-hoc tests with Holm--Bonferroni correction.
The significance level is set to $\alpha=0.05$.
In Figs.~\ref{fig:userstudy_boxplots_q2_v9} and~\ref{fig:userstudy_boxplots_q1_v9}, brackets labeled \emph{n.s.} denote method pairs that are not significantly different after correction.
These plots complement the aggregate scores reported in the main paper by explicitly showing response variability, overlap between methods, and statistically non-significant pairwise comparisons.

\newpage
\ifdefined\COMPLETIONANYFOURDMAIN
\else
\bibliographystyle{splncs04}
\bibliography{main}

\end{document}
\fi

\bibliography{main}
\end{document}